\documentclass{article}

\usepackage{microtype}
\usepackage{graphicx}
\usepackage{subcaption}
\usepackage{booktabs} % for professional tables

\usepackage{hyperref}

\usepackage[preprint]{icml2026}

\usepackage{amsmath}
\usepackage{amssymb}
\usepackage{mathtools}
\usepackage{amsthm}

\usepackage{url}
\usepackage{tikz}
\usepackage[utf8]{inputenc} %
\usepackage[T1]{fontenc}    %
\usepackage{amsfonts}       %
\usepackage{nicefrac}       %
\usepackage{xcolor}         %
\usepackage{float}
\usepackage{wrapfig}
\usepackage[capitalize,noabbrev]{cleveref}

\theoremstyle{plain}

\theoremstyle{definition}
\newtheorem{definition}{Definition}

\theoremstyle{remark}

\usepackage[textsize=tiny]{todonotes}

\icmltitlerunning{Direct Soft-Policy Sampling via Langevin Dynamics}

\begin{document}

\twocolumn[
  \icmltitle{Direct Soft-Policy Sampling via Langevin Dynamics}

  % It is OKAY to include author information, even for blind submissions: the
  % style file will automatically remove it for you unless you've provided
  % the [accepted] option to the icml2026 package.

  % List of affiliations: The first argument should be a (short) identifier you
  % will use later to specify author affiliations Academic affiliations
  % should list Department, University, City, Region, Country Industry
  % affiliations should list Company, City, Region, Country

  % You can specify symbols, otherwise they are numbered in order. Ideally, you
  % should not use this facility. Affiliations will be numbered in order of
  % appearance and this is the preferred way.
  \icmlsetsymbol{equal}{*}
  
  \begin{icmlauthorlist}
    \icmlauthor{Donghyeon Ki}{yyy}
    \icmlauthor{Hee-Jun Ahn}{yyy}
    \icmlauthor{Kyungyoon Kim}{yyy}
    \icmlauthor{Byung-Jun Lee}{yyy,comp}
    %\icmlauthor{}{sch}
    %\icmlauthor{}{sch}
  \end{icmlauthorlist}

  \icmlaffiliation{yyy}{Department of Artificial Intelligence, Korea University, Seoul, Republic of Korea}
  \icmlaffiliation{comp}{Gauss Labs Inc}

  \icmlcorrespondingauthor{Donghyeon Ki}{peop1e1n@korea.ac.kr}
  \icmlcorrespondingauthor{Byung-Jun Lee}{byungjunlee@korea.ac.kr}

  % You may provide any keywords that you find helpful for describing your
  % paper; these are used to populate the "keywords" metadata in the PDF but
  % will not be shown in the document
  % \icmlkeywords{Machine Learning, ICML}

  \vskip 0.3in
]

% this must go after the closing bracket ] following \twocolumn[ ...

% This command actually creates the footnote in the first column listing the
% affiliations and the copyright notice. The command takes one argument, which
% is text to display at the start of the footnote. The \icmlEqualContribution
% command is standard text for equal contribution. Remove it (just {}) if you
% do not need this facility.

% Use ONE of the following lines. DO NOT remove the command.
% If you have no special notice, KEEP empty braces:
\printAffiliationsAndNotice{Code is available at \url{https://github.com/ku-dmlab/NCLQL}.}  % no special notice (required even if empty)
% Or, if applicable, use the standard equal contribution text:
% \printAffiliationsAndNotice{\icmlEqualContribution}

\begin{abstract}
Soft policies in reinforcement learning define policies as Boltzmann distributions over state-action value functions, providing a principled mechanism for balancing exploration and exploitation. However, realizing such soft policies in practice remains challenging. Existing approaches either depend on parametric policies with limited expressivity or employ diffusion-based policies whose intractable likelihoods hinder reliable entropy estimation in soft policy objectives. We address this challenge by directly realizing soft-policy sampling via Langevin dynamics driven by the action gradient of the Q-function. This perspective leads to Langevin Q-Learning (LQL), which samples actions from the target Boltzmann distribution without explicitly parameterizing the policy. However, directly applying Langevin dynamics suffers from slow mixing in high-dimensional and non-convex Q-landscapes, limiting its practical effectiveness. To overcome this, we propose Noise-Conditioned Langevin Q-Learning (NC-LQL), which integrates multi-scale noise perturbations into the value function. NC-LQL learns a noise-conditioned Q-function that induces a sequence of progressively smoothed value landscapes, enabling sampling to transition from global exploration to precise mode refinement. On OpenAI Gym MuJoCo benchmarks, NC-LQL achieves competitive performance compared to state-of-the-art diffusion-based methods, providing a simple yet powerful solution for online RL.
\end{abstract}

\section{Introduction}
Recent advances in reinforcement learning (RL) have increasingly emphasized expressive policy representations that can capture complex, multimodal action distributions~\cite{wang2022diffusion, lu2023contrastive, park2025flow, ki2025prior}. In particular, soft policy frameworks~\citep{haarnoja2017reinforcement, haarnoja2018soft, haarnoja2018softactor, wang2024diffusion, dong2025maximum, ma2025efficient, celik2025dime} define policies as Boltzmann distributions over state-action value functions, providing a principled mechanism to balance exploration and exploitation by coupling value maximization with entropy regularization. However, realizing such policies in practice remains challenging. Conventional actor-critic methods rely on Gaussian policy parameterization whose limited expressivity often fails to represent multimodal Boltzmann distributions~\cite{haarnoja2018soft, haarnoja2018softactor}, while diffusion-based policies typically do not admit tractable policy densities, making entropy estimation difficult and often requiring costly approximations~\cite{wang2024diffusion, wang2025enhanced, dong2025maximum, celik2025dime}.

In this work, we revisit soft policy learning from a score-based perspective. A key observation is that the score function of soft policies $(\nabla_{\mathbf{a}}\log \pi_{\text{soft}}(\mathbf{a}|\mathbf{s}))$ depends solely on the action-gradient of the Q-function $(\nabla_{\mathbf{a}} Q(\mathbf{s}, \mathbf{a}))$. This property is particularly appealing because it implies that the target score function is directly available through policy evaluation, without requiring access to optimal action samples. This intrinsic characteristic of soft policies naturally aligns with Langevin dynamics, which requires only the score function to sample from a target probability density. Motivated by this connection, we propose \textit{Langevin Q-Learning} (LQL), an actor-free RL framework that directly samples actions from the Boltzmann distribution induced by the Q-function via Langevin dynamics. By replacing explicit policy parameterization with gradient-based sampling, LQL eliminates the need for separate actor updates, avoids entropy estimation altogether, and ensures that sampled actions are principled draws from the target soft policy.

Although theoretically attractive, LQL suffers from a critical practical limitation. Directly applying Langevin dynamics in high-dimensional and non-convex Q-landscapes can lead to slow mixing~\cite{song2019generative}. Because Langevin dynamics rely solely on local gradient information, sampling chains can become trapped in local modes and fail to explore the global structure of the action space. This issue is particularly severe in deep RL, where Q-functions typically exhibit highly rugged and multimodal geometries. 

To address this challenge, we introduce \textit{Noise-Conditioned Langevin Q-Learning} (NC-LQL). Our key idea is to incorporate multi-scale noise perturbations into the value function itself, yielding a noise-conditioned Q-function that induces a hierarchy of progressively smoothed value landscapes. This perspective is conceptually related to the noising mechanisms used in diffusion and score-based models, but differs in that noise is introduced directly at the level of the value function rather than the policy density. At large noise scales, smoothing bridges isolated modes and facilitates global exploration; as the noise is annealed, the value landscape gradually recovers fine-grained structure, enabling precise local refinement. By sampling actions through annealed Langevin dynamics guided by this noise-conditioned Q-function, NC-LQL enables efficient mixing while faithfully recovering the target Boltzmann distribution. 
Empirically, NC-LQL achieves competitive or superior performance on OpenAI Gym MuJoCo benchmarks compared to state-of-the-art diffusion-based online RL methods, while using fewer parameters and a significantly simpler learning pipeline.
By directly sampling from the Boltzmann distribution via score-based dynamics, our approach eliminates the need for an explicit actor and provides a simple yet powerful alternative to existing diffusion-based methods.

\section{Preliminaries}
\label{sec:preliminaries}
\paragraph{Reinforcement Learning (RL)} We consider the RL problem under a Markov Decision Process (MDP) $\mathcal{M}=\{\mathcal{S},\mathcal{A},\mathcal{P},r,p_0,\gamma\}$, where $\mathcal{S}$ is the state space, $\mathcal{A}$ is the action space, $\mathcal{P}(\mathbf{s}'| \mathbf{s},\mathbf{a})$ is the transition probability, $r(\mathbf{s}, \mathbf{a})$ is the reward, $p_0(\mathbf{s})$ is the initial-state distribution and $\gamma\in[0,1)$ is the discount factor. The goal of RL is to learn a policy that maximizes the expected cumulative discounted reward~\cite{sutton1998reinforcement,konda1999actor}. For a policy $\pi(\mathbf{a} | \mathbf{s})$, the state-action value is defined as $Q^\pi(\mathbf{s},\mathbf{a})=\mathbb{E} \left[\sum_{\tau=0}^{\infty}\gamma^\tau r(\mathbf{s}_\tau,\mathbf{a}_\tau)|\mathbf{s}_0=\mathbf{s},\mathbf{a}_0=\mathbf{a},\pi,\mathcal{P}\right]$, where $\tau$ denotes the environment timesteps. A central principle of RL is the policy iteration framework, which alternates between two steps. First, \textit{policy evaluation} estimates $Q^\pi$ for a fixed policy, typically by iterating the Bellman operator:
\begin{align}
\label{eq:bellman_operator}
    \mathcal{T}^\pi Q(\mathbf{s},\mathbf{a}) := r(\mathbf{s},\mathbf{a}) + \gamma\,\mathbb{E}_{\mathbf{s}' \sim \mathcal{P}, \mathbf{a}'\sim \pi(\cdot|\mathbf{s}')} [Q(\mathbf{s}',\mathbf{a}')]
\end{align}
Second, \textit{policy improvement} updates $\pi$ towards actions that maximize the expected Q-value, e.g., $\pi(\cdot|\mathbf{s}) \leftarrow \arg\max_{\pi} \mathbb{E}_{\mathbf{a} \sim \pi} [Q^\pi(\mathbf{s},\mathbf{a})]$. Actor-critic methods instantiate this paradigm in a parametric form: the critic approximates $Q^\pi$ through Bellman backups, while the actor is updated using the critic’s value estimates, thus coupling policy evaluation and improvement in a single learning loop.

\paragraph{Langevin Dynamics}
Langevin dynamics provides a mechanism to sample from a probability density $p(\mathbf{x})$ by utilizing the score function, $\nabla_\mathbf{x} \log p(\mathbf{x})$. Given a step size $\epsilon >0$, a total number of iterations $T$ and an initial sample $\mathbf{x}_0$, Langevin dynamics iteratively computes the following:
\begin{align*}
    \mathbf{x}_{t} = \mathbf{x}_{t-1} + \frac{\epsilon}{2} \nabla_{\mathbf{x}}\log p(\mathbf{x}_{t-1}) + \sqrt{\epsilon} \mathbf{z}_t, \quad  \text{for}~~t \leftarrow 1 ~\text{to}~ T 
\end{align*}
where $\mathbf{z}_t \sim \mathcal{N}(\mathbf{0},\mathbf{I})$. As $\epsilon \to 0$ and $T \to \infty$, the distribution of $\mathbf{x}_T$ converges to $p(\mathbf{x})$ under some regularity conditions~\cite{roberts1996exponential,welling2011bayesian}. However, in generative modeling tasks, the analytic form of the score function $\nabla_\mathbf{x} \log p(\mathbf{x})$ is unknown; instead, only empirical samples from $p(\mathbf{x})$ are available. To address this, a neural network $s_\theta(\mathbf{x})$, known as a score network, is optimized using these samples to approximate the score function, typically via score matching objectives.

\paragraph{Score-Based Generative Modeling}

\begin{algorithm}[t!]
\caption{Annealed Langevin Dynamics (ALD)}
\label{alg:ald}
\begin{algorithmic}
\STATE {\bfseries Input:} score network $s_\theta (\tilde{\mathbf{x}},\sigma_i)$, total number of iterations $T$, step size $\epsilon$, noise scales $\{\sigma_i\}^L_{i=1}$, initial sample $\mathbf{x}_{0}$
\STATE $\mathbf{x}_{1,0} \leftarrow \mathbf{x}_0$
\FOR{$i \leftarrow 1$ to $L$}
\STATE $\alpha_i \leftarrow \epsilon \cdot \sigma^2_i / \sigma^2_L$
\FOR{$t \leftarrow 1$ to $T$}
\STATE Sample $\mathbf{z}_t \sim \mathcal{N}(\mathbf{0}, \mathbf{I})$
\STATE $\mathbf{x}_{i,t} \leftarrow \mathbf{x}_{i,t-1} + \frac{\alpha_i}{2}s_\theta(\mathbf{x}_{i,t-1}, \sigma_i) + \sqrt{\alpha_i}\mathbf{z}_t$
\ENDFOR
\STATE $\mathbf{x}_{i+1,0} \leftarrow \mathbf{x}_{i,T}$
\ENDFOR
\STATE \textbf{Return} $\mathbf{x}_{L,T}$ 
\end{algorithmic}
% \vspace{-0.5cm}
\end{algorithm}

\begin{figure*}[ht!]
    \centering
    \includegraphics[width=1.0\linewidth]{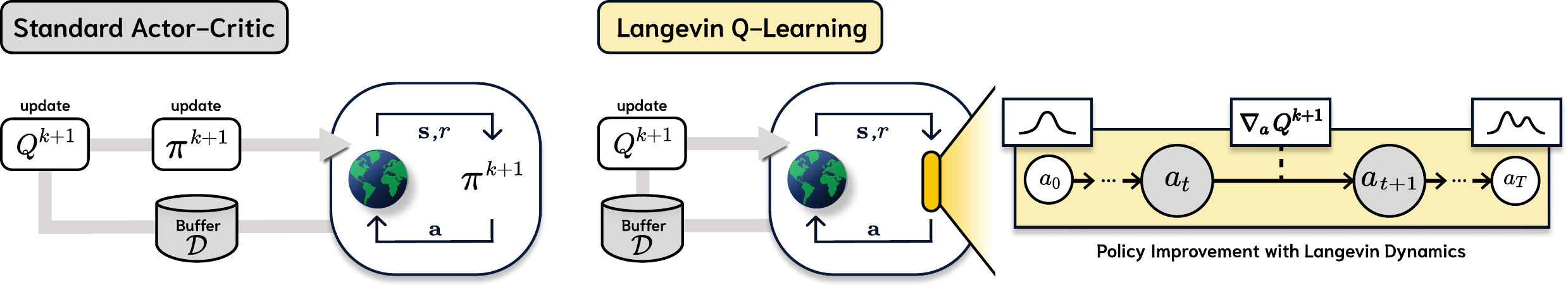}
    \caption{Comparison between standard actor-critic and Langevin Q-Learning (LQL). While standard actor-critic methods rely on explicit actor updates to approximate a target policy, LQL directly samples actions from the Boltzmann distribution defined by the Q-function via Langevin dynamics, removing the need for a separate actor update.}
    \vspace{-0.3cm}
\end{figure*}

Score-based generative modeling~\cite{song2019generative} enables sampling by estimating the score function of the data distribution. However, the estimated score function is often inaccurate in regions without training data, causing Langevin dynamics to struggle with mixing and frequently become trapped in local modes. To address this, \citet{song2019generative} perturbs the data with multiple levels of Gaussian noise, producing a sequence of noise-perturbed distributions. This perturbation ensures the resulting distributions do not collapse onto a low-dimensional manifold, thereby maintaining well-defined gradients over the entire space. Specifically, a sequence of noise scales $\{\sigma_i\}_{i=1}^L$ is defined such that $\sigma_1 > \cdots > \sigma_L > 0$. This yields a sequence of noise-perturbed distributions $q_{\sigma_i}(\tilde{\mathbf{x}}) = \int p(\mathbf{x})\mathcal{N}(\tilde{\mathbf{x}}|\mathbf{x},\sigma_i^2\mathbf{I})d\mathbf{x}$. To estimate the scores across all noise scales $\nabla_{\tilde{\mathbf{x}}} \log q_{\sigma_i}(\tilde{\mathbf{x}})$, a single conditional score network $s_\theta(\tilde{\mathbf{x}}, \sigma_i)$ is trained by minimizing the Denoising Score Matching (DSM) loss~\cite{vincent2011connection} across all noise levels:
\begin{align}
\label{eq:dsm}
    \sum_{i=1}^L \lambda(\sigma_i) \mathbb{E}_{p(\mathbf{x})}\mathbb{E}_{q_{\sigma_i}(\tilde{\mathbf{x}}|\mathbf{x})} \left[ \left\| s_\theta(\tilde{\mathbf{x}}, \sigma_i) + \frac{\tilde{\mathbf{x}} - \mathbf{x}}{\sigma_i^2} \right\|_2^2 \right].
\end{align}
where $\lambda(\sigma_i)$ is a weighting function. Sampling is performed via Annealed Langevin Dynamics (ALD), which iteratively refines an initial random vector across the noise sequence. The overall process of ALD is detailed in \cref{alg:ald}. As the final noise scale $\sigma_L \approx 0$, and provided that the Langevin dynamics converges, the distribution of the resulting samples $\mathbf{x}_{L,T}$ converges to $p(\mathbf{x})$.

\section{Langevin Q-Learning}
\label{sec:langevin_qlearning}
In online RL, soft policies provide a principled mechanism for balancing action value maximization and stochastic exploration. Such policies are commonly defined as Boltzmann distributions over the state-action value function~\citep{haarnoja2018soft, wang2024diffusion}:
\begin{align}
    \pi_\text{soft}(\mathbf{a}|\mathbf{s}) = \exp\left(Q(\mathbf{s},\mathbf{a})\right) / Z(\mathbf{s})
\end{align}
where $Z(\mathbf{s}) = \int \exp\left(Q(\mathbf{s},\mathbf{a})\right)d\mathbf{a}$. Learning soft policies can be formulated as minimizing the KL-divergence between the policy and the target Boltzmann distribution $\mathcal{D}_{\text{KL}}\left(\pi(\cdot|\mathbf{s}) \big\Vert \frac{\exp(Q(\mathbf{s},\cdot))}{Z(\mathbf{s})}\right)$. This objective admits an equivalent form as maximizing an entropy-regularized objective:
\begin{align}
\max_\pi~\mathbb{E}_{\mathbf{s} \sim \mathcal{B}, \mathbf{a}\sim \pi}
\left[ Q(\mathbf{s},\mathbf{a}) + \mathcal{H}(\pi(\cdot|\mathbf{s})) \right].
\label{eq:policy_improvement}
\end{align}
where $\mathcal{B}$ is a replay buffer. Despite the theoretical appeal of soft policies, previous approaches encounter two key challenges. First, Gaussian parametric policies often lack the expressivity required to accurately represent complex, potentially multimodal Boltzmann distributions. Second, while diffusion-based policies offer higher expressivity, their induced action distributions typically do not admit a tractable log-density. As a result, entropy terms are often estimated through complex or computationally expensive approximations, which makes it difficult to guarantee that the resulting samples correspond exactly to the Boltzmann distribution.

To ensure that samples are accurately drawn from the target Boltzmann distribution, we note a fundamental property of soft policies: its score function depends solely on the action-gradient of the Q-function:
\begin{align}
\nabla_{\mathbf{a}} \log \pi_{\text{soft}}(\mathbf{a}|\mathbf{s}) = \nabla_{\mathbf{a}} Q(\mathbf{s},\mathbf{a}).
\end{align}
This property highlights a crucial distinction between RL and supervised score-based generative modeling. In supervised learning, the score must be estimated from a dataset of ground-truth samples~\cite{song2019generative, song2020score, song2021maximum, lai2025principles}. However, in RL, although we lack action samples from the target distribution, policy evaluation provides us with the Q-function, giving us direct access to the target score function. 

This intrinsic characteristic of soft policies aligns perfectly with Langevin dynamics, which requires only the score function to sample from a target probability density. By leveraging this, we can sample actions from the soft policy by only using the action-gradient of the Q-function.
\begin{definition}[\textbf{Langevin soft policy}] 
\label{def:langevin_improvement}
Given the current Q-function $Q(\mathbf{s},\mathbf{a})$ and a state $\mathbf{s}$, the \textit{Langevin soft policy} $\pi_{\text{LD}}$ is defined as the distribution induced by the following Langevin dynamics sampling procedure. We initialize $\mathbf{a}_0$ and apply the update sequentially for $t \leftarrow 1$ to $T$:
\begin{align*}
    \mathbf{a}_{t}
    = \mathbf{a}_{t-1}
      + \frac{\epsilon}{2}\nabla_{\mathbf{a}} Q(\mathbf{s}, \mathbf{a}_{t-1})
      + \sqrt{\epsilon}\,\mathbf{z}_{t},
    \quad \mathbf{z}_{t} \sim \mathcal{N}(\mathbf{0}, \mathbf{I}).
\end{align*}
The resulting action $\mathbf{a}_T$ follows the policy $\pi_{\text{LD}}(\cdot | \mathbf{s})$.
\end{definition}
As the step size $\epsilon \to 0$ and the number of steps $T \to \infty$, the distribution of $\mathbf{a}_T$ converges to the Boltzmann distribution $\exp(Q(\mathbf{s}, \mathbf{a})) / Z(\mathbf{s})$. Importantly, the resulting policy $\pi_{\text{LD}}$ is obtained without introducing any explicit policy parameters or performing a separate policy optimization step.

By leveraging the Langevin soft policy, we propose \textit{Langevin Q-Learning} (LQL), a framework that eliminates the need for an explicit parametric policy. Instead of learning a separate policy network, LQL generates actions directly through Langevin dynamics driven by the gradient of the Q-function. This design enables principled sampling from the exact Boltzmann distribution induced by the current value function, without requiring tractable policy densities or entropy approximations.

\begin{figure*}[t]
    \centering
    \includegraphics[width=1.0\linewidth]{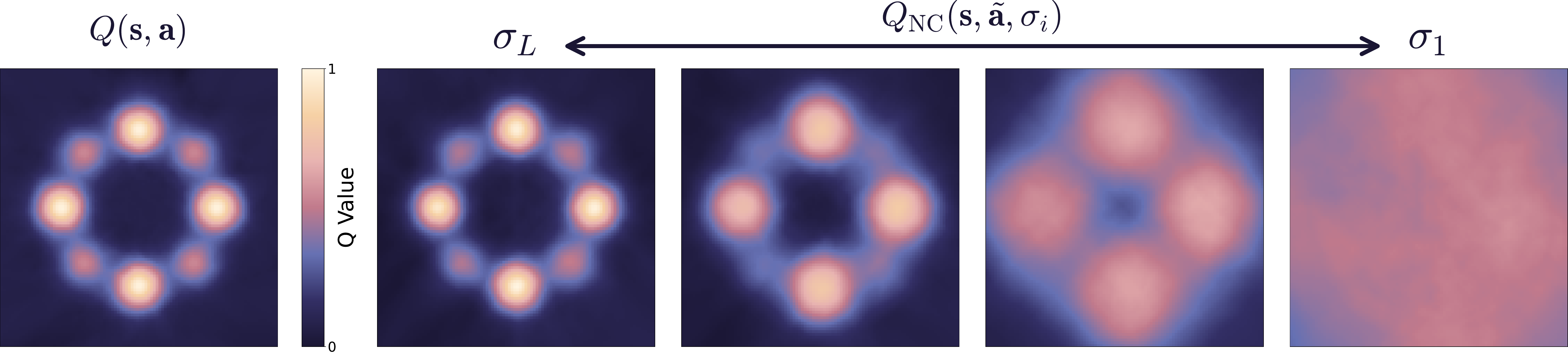}
    \caption{Visualization of value maps in a 2D bandit environment. We show the noise-conditioned Q-function $Q_{\text{NC}}(\mathbf{s}, \tilde{\mathbf{a}}, \sigma_i)$ at different noise scales ($i=1,4,7,10$), compared with the standard Bellman critic $Q(\mathbf{s}, \mathbf{a})$. Experimental details are provided in Appendix~\ref{appendix:bandit_env}.}
    \label{fig:noise-conditioned_Q}
    \vspace{-0.2cm}
\end{figure*}

\paragraph{Langevin Q-Learning} LQL approximates the state-action value function $Q_\theta$ using a neural network and minimizes a standard temporal difference (TD) error over transitions sampled from a replay buffer $\mathcal{B}$:
\begin{align}
\label{eq:lql_objective}
    &\min_\theta~\mathbb{E}_{\mathbf{s}, \mathbf{a}} \left[ \left( Q_\theta(\mathbf{s}, \mathbf{a}) - \mathcal{T}^{\pi_{\text{LD}}} Q_{\bar{\theta}}(\mathbf{s},\mathbf{a}) \right)^2 \right], \\
    &\text{where}~~ \mathcal{T}^{\pi_{\text{LD}}}Q_{\bar{\theta}} = r + \gamma \mathbb{E}_{\mathbf{s}'\sim \mathcal{B}, \mathbf{a}' \sim \pi_{\text{LD}}(\cdot| \mathbf{s}')} \left[Q_{\bar{\theta}}(\mathbf{s}', \mathbf{a}')\right], \nonumber
\end{align}
Here, $r = r(\mathbf{s},\mathbf{a})$ and $Q_{\bar{\theta}}$ represents the target network. $\pi_{\text{LD}}$ denotes the \textit{Langevin soft policy} defined in \cref{def:langevin_improvement}. Previous actor-critic methods often alternate between updating the Q-function via the Bellman equation and explicitly optimizing the policy $\pi$ by maximizing the objective in \cref{eq:policy_improvement} to realize sampling from the Boltzmann distribution. In contrast, LQL simplifies this iterative process: it directly realizes Boltzmann sampling through the Langevin soft policy, where actions $\mathbf{a}'$ used in the Bellman target are generated via Langevin dynamics driven by the Q-function gradient. As a result, learning in LQL reduces to optimizing the Q-function alone, driving it toward optimality without the need to maintain or update a separate policy network. The overall algorithm is detailed in Appendix~\ref{appendix:lql}.

\section{Noise-Conditioned Langevin Q-Learning}
While LQL provides a theoretically sound mechanism for sampling from the soft policy without an explicit actor, its practical effectiveness is constrained by the slow mixing rates of Langevin dynamics~\cite{song2019generative}. Because the dynamics are driven solely by local score information ($\nabla_\mathbf{a} Q(\mathbf{s}, \mathbf{a})$), they struggle to navigate the highly non-convex and multimodal Q-landscapes typical of deep reinforcement learning. In such landscapes, sampling chains frequently become trapped in local optima, unable to cross high-energy barriers to reach the global optimum.

To overcome these challenges, we propose \textit{Noise-Conditioned Langevin Q-Learning} (NC-LQL). This approach introduces a multi-scale noise perturbation mechanism that generates smoothed Q-landscapes. By annealing the noise scale, we create a navigable path through the non-convex value surface, effectively guiding the sampling dynamics toward the target Boltzmann distribution.

\subsection{Multi-Scale Noise Perturbation}

We begin by formalizing the multi-scale noise perturbation mechanism underlying NC-LQL. Specifically, we define a sequence of noise scales $\{\sigma_{i}\}_{i=1}^{L}$ such that $\sigma_{1}>\dots>\sigma_{L} \approx 0$. At each noise scale $\sigma_i$, we consider a noise injection process where a noisy action $\tilde{\mathbf{a}}$ is generated from a clean action $\mathbf{a}$ according to the Gaussian distribution $\mathcal{N}(\tilde{\mathbf{a}}|\mathbf{a}, \sigma_i^2 \mathbf{I})$. Given the $Q$-function obtained from standard policy evaluation~(by~\cref{eq:bellman_operator}) , we define the \textit{noise-conditioned Q-function}, $Q_{\text{NC}}$, as the conditional expectation of $Q$ under the posterior distribution:
\begin{align}
\label{eq:q_ec_def}
    Q_{\text{NC}}(\mathbf{s}, \tilde{\mathbf{a}}, \sigma_i) := \mathbb{E}_{\mathbf{a} \sim p(\cdot|\tilde{\mathbf{a}}, \mathbf{s}, \sigma_i)} [Q(\mathbf{s}, \mathbf{a})]
\end{align}
where $p(\mathbf{a}|\tilde{\mathbf{a}}, \mathbf{s}, \sigma_i) \propto \mathcal{N}(\tilde{\mathbf{a}}|\mathbf{a}, \sigma_i^2 \mathbf{I}) p(\mathbf{a}|\mathbf{s})$ denotes the posterior distribution of the clean action $\mathbf{a}$ given the noisy action $\tilde{\mathbf{a}}$. Here, $p(\mathbf{a}|\mathbf{s})$ is the prior distribution over actions, taken in practice as the empirical distribution of the replay buffer.

Intuitively, for a fixed noisy action $\tilde{\mathbf{a}}$, this definition aggregates the original Q-values of all clean actions $\mathbf{a}$ that could generate $\tilde{\mathbf{a}}$ under the noising process $\mathcal{N}(\tilde{\mathbf{a}} | \mathbf{a}, \sigma_i^2 \mathbf{I})$. In this sense, $Q_{\text{NC}}$ can be interpreted as an averaged value function that smooths the original Q-landscape according to the noise scale $\sigma_i$. At large noise scales (e.g., $\sigma_1$), this smoothing bridges separated local modes and reshapes the value landscape, allowing the dynamics to traverse high-energy barriers and explore the global structure of the action space. As the noise scale anneals toward $\sigma_L$, the smoothing effect gradually diminishes, and the value landscape recovers its fine-grained complexity. In the limit $\sigma_L \approx 0$, the noising process becomes negligible, and the noise-conditioned Q-function reduces to the original value function, i.e., $Q_{\text{NC}}(\cdot, \cdot, \sigma_L) \approx Q(\cdot, \cdot)$. Consequently, the dynamics transition from global exploration to precise local refinement and converge to sharp modes. This multi-scale behavior mitigates the slow-mixing issue inherent in Langevin dynamics. Visualizations of the noise-conditioned Q-function are provided in \cref{fig:noise-conditioned_Q}.

Building on this smoothed value representation, we define a policy that samples actions via annealed Langevin dynamics driven by the action-gradient of the noise-conditioned Q-function, $\nabla_{\mathbf{\tilde{a}}}Q_{\text{NC}}(\mathbf{s},\mathbf{\tilde{a}},\sigma_i)$. 

\begin{definition}[\textbf{Noise-conditioned Langevin soft policy}]
\label{def:ec_langevin_improvement}
Given the current Q-function $Q(\mathbf{s},\mathbf{a})$, a state $\mathbf{s}$, and a sequence of noise scales $\sigma_1 > \sigma_2 > \dots > \sigma_L > 0$, the \textit{noise-conditioned Langevin soft policy} $\pi_{\text{NC}}$ is defined as the distribution induced by the following annealed Langevin dynamics sampling procedure. We initialize $\mathbf{a}_{1,0}$ with $\mathbf{a}_0$. For $i \leftarrow 1$ to $L$, we perform $T$ steps of Langevin dynamics using the noise-conditioned Q-function $Q_{\text{NC}}(\cdot, \cdot, \sigma_i)$. Specifically, for $t \leftarrow 1$ to $T$, the update rule is given by
\begin{align*}
    \mathbf{a}_{i,t}
    = \mathbf{a}_{i,t-1}
    + \frac{\alpha_i}{2}\nabla_{\tilde{\mathbf{a}}} Q_{\text{NC}}(\mathbf{s}, \mathbf{a}_{i,t-1}, \sigma_i)
    + \sqrt{\alpha_i}\,\mathbf{z}_t,
\end{align*}
where $\alpha_i$ is a step size adapted to the noise level and $\mathbf{z}_t \sim \mathcal{N}(\mathbf{0}, \mathbf{I})$. 
After completing the inner loop, the final sample at noise level $i$ is used to initialize the next level, i.e., $\mathbf{a}_{i+1,0} \leftarrow \mathbf{a}_{i,T}$. 
The resulting action $\mathbf{a}_{L,T}$ follows the policy $\pi_{\text{NC}}(\cdot \mid \mathbf{s})$. 
The detailed pseudo-code is provided in Appendix~\ref{appendix:noise-conditioned langevin soft policy}.
\end{definition}

As the noise scale approaches the limit $\sigma_{L}\approx 0$, the gradient of the smoothed value landscape $\nabla_{\tilde{\mathbf{a}}} Q_{\text{NC}}(\cdot, \cdot, \sigma_L)$ converges to the target score function $\nabla_\mathbf{a} Q(\cdot, \cdot)$. Thus, consistent with the convergence of annealed Langevin dynamics~\cite{song2019generative}, the final sample $\mathbf{a}_{L,T}$ approximates a draw from the Boltzmann distribution $\exp(Q(\mathbf{s},\mathbf{a}))/Z(\mathbf{s})$.

\subsection{Noise-Conditioned Langevin Q-Learning}

By extending LQL with noise-conditioned value representations, we now introduce a \textit{Noise-Conditioned Langevin Q-Learning} (NC-LQL). We utilize a single neural network $Q_{\theta}(\mathbf{s},\tilde{\mathbf{a}},\sigma_i)$ to simultaneously approximate the standard Bellman Q-function and its smoothed variants across all noise scales. The training objective minimizes a joint loss over transitions sampled from a replay buffer $\mathcal{B}$:
\begin{align}
    &\min_\theta ~\mathbb{E}_{\mathbf{s},\mathbf{a}}\left[\left(Q_\theta(\mathbf{s},\mathbf{a}, \sigma_L) - \mathcal{T}^{\pi_{\text{NC}}} Q_{\bar{\theta}} (\mathbf{s},\mathbf{a}, \sigma_L)\right)^2\right] \label{eq:ec_bellman}\\
    &+ \mathbb{E}_{\mathbf{s},\mathbf{a}, \tilde{\mathbf{a}}, i}\left[\left(Q_\theta(\mathbf{s},\tilde{\mathbf{a}}, \sigma_i) - \texttt{sg}\left(Q_\theta(\mathbf{s},\mathbf{a},\sigma_L)\right)\right)^2\right] \label{eq:ec_smooth}
\end{align}
\vspace{-0.5cm}
\begin{align*}
    \text{where}~~\mathcal{T}^{\pi_{\text{NC}}} Q_{\bar{\theta}} = r +\gamma \mathbb{E}_{\mathbf{s}' \sim \mathcal{B}, \mathbf{a}'\sim \pi_{\text{NC}}(\cdot|\mathbf{s}')}\left[Q_{\bar{\theta}}(\mathbf{s}',\mathbf{a}', \sigma_L)\right].
\end{align*}
Here, $\tilde{\mathbf{a}} \sim \mathcal{N}(\cdot|\mathbf{a},\sigma_i^2\mathbf{I})$, and $i \sim \mathcal{U}[1,L]$. In this objective, $\pi_{\text{NC}}$ denotes the \textit{noise-conditioned Langevin soft policy} defined in~\cref{def:ec_langevin_improvement}. Note that NC-LQL preserves the actor-free structure of LQL, in which learning is driven solely by noise-conditioned Q-function optimization rather than explicit policy updates. The overall procedure is described in~\cref{alg:eclql}. In the following, we analyze \cref{eq:ec_bellman,eq:ec_smooth} to clarify the role of each term.

\begin{algorithm}[tb]
\caption{Noise-Conditioned Langevin Q-Learning}
\label{alg:eclql}
\begin{algorithmic}
\STATE {\bfseries Input:} $\mathcal{B}$, $Q_\theta(\mathbf{s},\tilde{\mathbf{a}}, \sigma_i)$, $T$, $\epsilon$, $\{\sigma_i\}^L_{i=1}$
\FOR{each iteration}
\FOR{each sampling step}
\STATE Sample $\mathbf{a} \sim \pi_{\text{NC}}(\cdot|\mathbf{s})$ by \cref{def:ec_langevin_improvement}
\STATE Execute $\mathbf{a}$, observe reward $r$ and next state $\mathbf{s}'$
\STATE Store transition $(\mathbf{s}, \mathbf{a}, r, \mathbf{s}')$ in buffer $\mathcal{B}$
\ENDFOR
\FOR{each update step}
\STATE Sample mini-batch from $\mathcal{B}$
\STATE Update critic $Q_\theta$ with \cref{eq:ec_bellman} and \eqref{eq:ec_smooth}
\ENDFOR
\ENDFOR
\end{algorithmic}
\end{algorithm}

\begin{figure*}
    \centering
    \includegraphics[width=\linewidth]{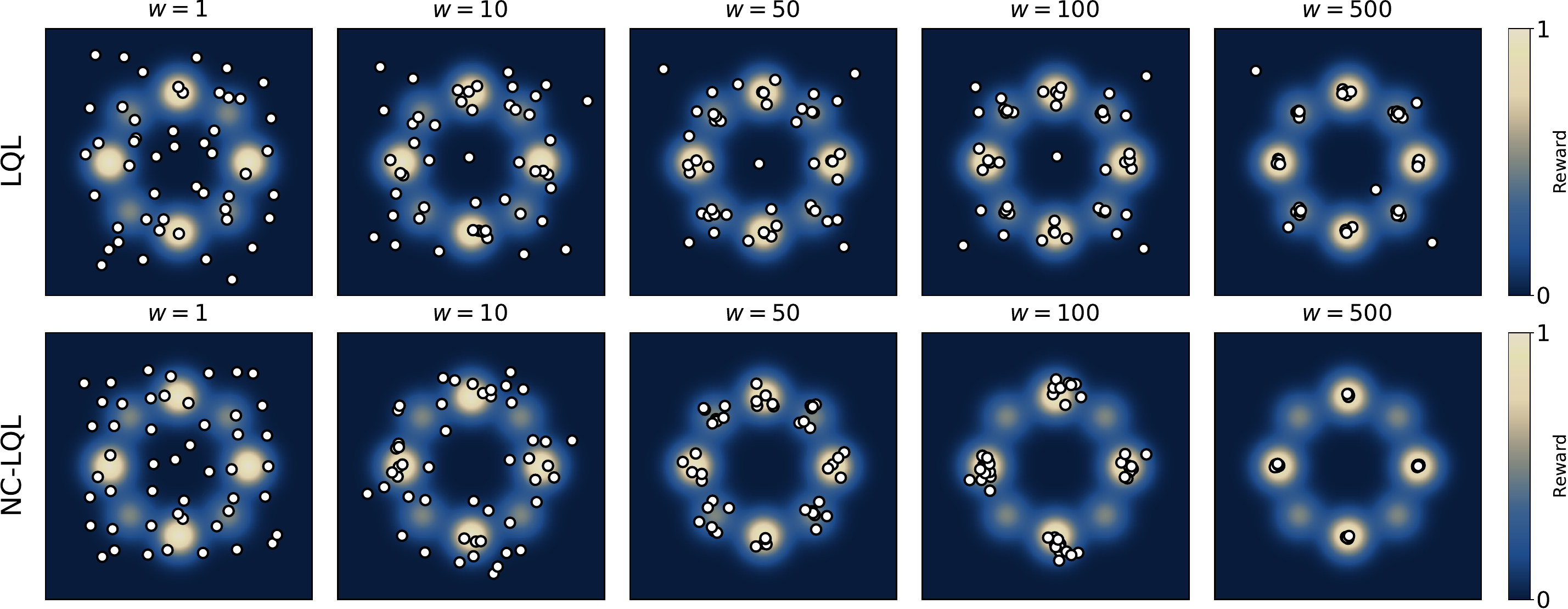}
    \caption{Visualization of samples obtained across different temperature parameters $w$ in the 2D bandit environment. The background represents the ground-truth reward landscape. Detailed experimental setup is provided in Appendix~\ref{appendix:bandit_env}.}
    \label{fig:bandit_w}
    \vspace{-0.2cm}
\end{figure*}

\paragraph{TD-learning at $\sigma_L$} \cref{eq:ec_bellman} enforces standard temporal difference learning at the minimum noise level $\sigma_L \approx 0$. This term ensures that $Q_{\text{NC}}(\cdot,\cdot,\sigma_L)$ converges pointwise to the standard Bellman Q-function $Q(\cdot,\cdot)$. 

\paragraph{Value Smoothing}
\cref{eq:ec_smooth} is a simple regression loss, yet it naturally induces smoothing landscapes required for efficient sampling. This follows from the fact that the minimizer of a mean squared error objective corresponds to a conditional expectation. Concretely, by regressing the noisy value $Q_\theta(\mathbf{s}, \tilde{\mathbf{a}}, \sigma_i)$ toward the clean target $Q_\theta(\mathbf{s}, \mathbf{a}, \sigma_L)$, the network is encouraged to learn
\begin{align*}
Q_{\text{NC}}(\mathbf{s}, \tilde{\mathbf{a}}, \sigma_i) &= \mathbb{E}_\mathbf{a}\left[Q_{\text{NC}}(\mathbf{s},\mathbf{a},\sigma_L)|\tilde{\mathbf{a}}\right] \quad~\text{(by Eq.~\eqref{eq:ec_smooth})}\\ 
&\approx \mathbb{E}_\mathbf{a}\left[ Q(\mathbf{s}, \mathbf{a}) | \tilde{\mathbf{a}} \right].\quad~~~~~~~~~~\text{(by Eq.~\eqref{eq:ec_bellman})}
\end{align*}
which coincides with the definition of the noise-conditioned Q-function in \cref{eq:q_ec_def}. As a result, the value network implicitly learns smoothed Q-landscapes across noise scales, enabling isolated modes to be connected through a simple regression objective.

\paragraph{Relation to Previous Annealing Method}
In score-based generative modeling, slow mixing is addressed by constructing a sequence of noise-perturbed distributions that progressively smooth the underlying density~\citep{song2019generative}. Specifically, a sequence of noise-perturbed distributions $q_{\sigma_i}(\tilde{\mathbf{x}}) = \int p(\mathbf{x})\mathcal{N}(\tilde{\mathbf{x}}|\mathbf{x}, \sigma_i^2 \mathbf{I})d\mathbf{x}$ is defined by convolving the target distribution with Gaussian noise at multiple scales, which smooths the energy landscape and bridges isolated modes. However, in reinforcement learning settings where the policy is formulated as a Boltzmann distribution $\pi \propto \exp(Q)$, this approach requires estimating the score of the noise-perturbed policy $\pi_{\sigma_i}(\tilde{\mathbf{a}}|\mathbf{s})$:
\begin{align*}
    \pi_{\sigma_i}(\tilde{\mathbf{a}}|\mathbf{s}) &= \int\frac{\exp(Q(\mathbf{s},\mathbf{a}))}{Z(\mathbf{s})}\mathcal{N}(\tilde{\mathbf{a}}|\mathbf{a},\sigma^2_i \mathbf{I})d\mathbf{a}\\
    \nabla_{\tilde{\mathbf{a}}}\log \pi_{\sigma_i}(\tilde{\mathbf{a}}|\mathbf{s}) &= \nabla_{\tilde{\mathbf{a}}} \log \int \exp(Q(\mathbf{s}, \mathbf{a})) \mathcal{N}(\tilde{\mathbf{a}}|\mathbf{a}, \sigma^2_i \mathbf{I}) d\mathbf{a}.
\end{align*}
While \citet{song2019generative} employs Denoising Score Matching loss (\cref{eq:dsm}) to estimate the noise-perturbed score using ground-truth samples, this method is inapplicable in RL due to the lack of optimal samples. This necessitates a direct computation of the score, which introduces a severe bottleneck as it requires a log-sum-exp integration over the entire action space. Consequently, these computational challenges make the direct application of density perturbation in reinforcement learning settings prohibitive. 

By introducing multi-scale noise perturbations directly into the value function, NC-LQL reshapes the sampling geometry to mitigate the slow-mixing behavior of Langevin dynamics, while retaining principled Boltzmann sampling and the actor-free formulation of LQL. Although this approach perturbs the value landscape rather than the policy density, it serves an analogous role to density noising in diffusion-based methods by smoothing the underlying energy surface and enabling sampling trajectories to traverse otherwise prohibitive high-energy barriers. In contrast to the previous annealing approach, this design avoids intractable score estimation and costly log-sum-exp computations, making it directly applicable in reinforcement learning settings.

\section{Implementation Details}
\subsection{Soft Policy Temperature}
In soft policy frameworks, a temperature parameter is commonly introduced to balance value maximization and entropy regularization. Specifically, the soft policy is defined as $\pi \propto \exp\left(w Q\right)$, where $w > 0$ controls the trade-off between expected return and policy entropy. This policy is equivalent to the optimizer of the following objective $J(\pi) = \mathbb{E}_{\mathbf{s} \sim \mathcal{B}, \mathbf{a}\sim \pi} \left[Q(\mathbf{s}, \mathbf{a}) + \frac{1}{w} \mathcal{H}(\pi(\cdot \mid \mathbf{s}))\right]$. Under this formulation, the score function of the target soft policy is given by $\nabla_{\mathbf{a}}\log \pi (\mathbf{a}|\mathbf{s}) = w\nabla_{\mathbf{a}} Q(\mathbf{s}, \mathbf{a})$. We adopt the same temperature parameter in our method to balance the trade-off between exploitation and exploration.

To analyze the effect of the temperature parameter $w$, we visualize action samples produced by LQL and NC-LQL in a 2D bandit environment across various $w$s, as shown in \Cref{fig:bandit_w}. As $w$ increases, the target Boltzmann distribution assigns higher probability mass to high-reward regions, leading to sharper concentration around the highest-reward modes. LQL, however, suffers from slow mixing and exhibits little qualitative change as $w$ increases, with samples remaining dispersed or trapped around suboptimal local modes even at high temperatures. In contrast, NC-LQL mitigates this slow-mixing behavior and effectively captures the anticipated temperature-dependent behavior, producing samples that gradually sharpen with increasing $w$ and reflect the intended change in the target Boltzmann distribution.

\subsection{Algorithmic Simplicity}
Notably, our method \textbf{does not rely on auxiliary sampling heuristics} commonly used in recent diffusion-based approaches~\cite{psenka2023learning, ding2024diffusion, ma2025efficient}, such as best-of-$N$ action selection or injecting additional noise into sampled actions. Action sampling is performed directly from the soft policy, highlighting the simplicity and efficiency of our approach. 

\begin{figure*}[t!]
    \centering
    \includegraphics[width=0.3671\linewidth]{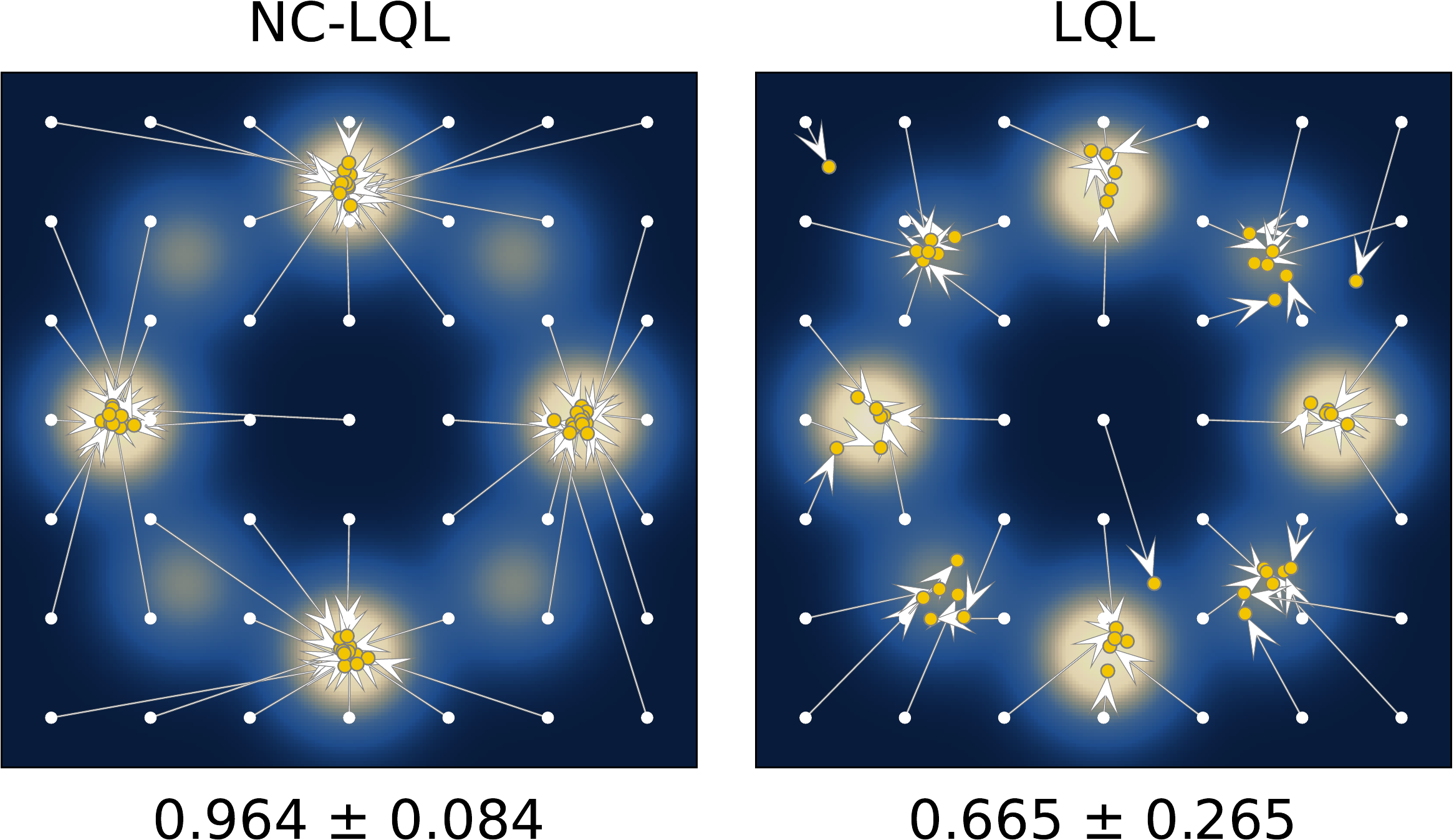}
    \hspace{0.066cm}
    \includegraphics[width=0.56\linewidth]{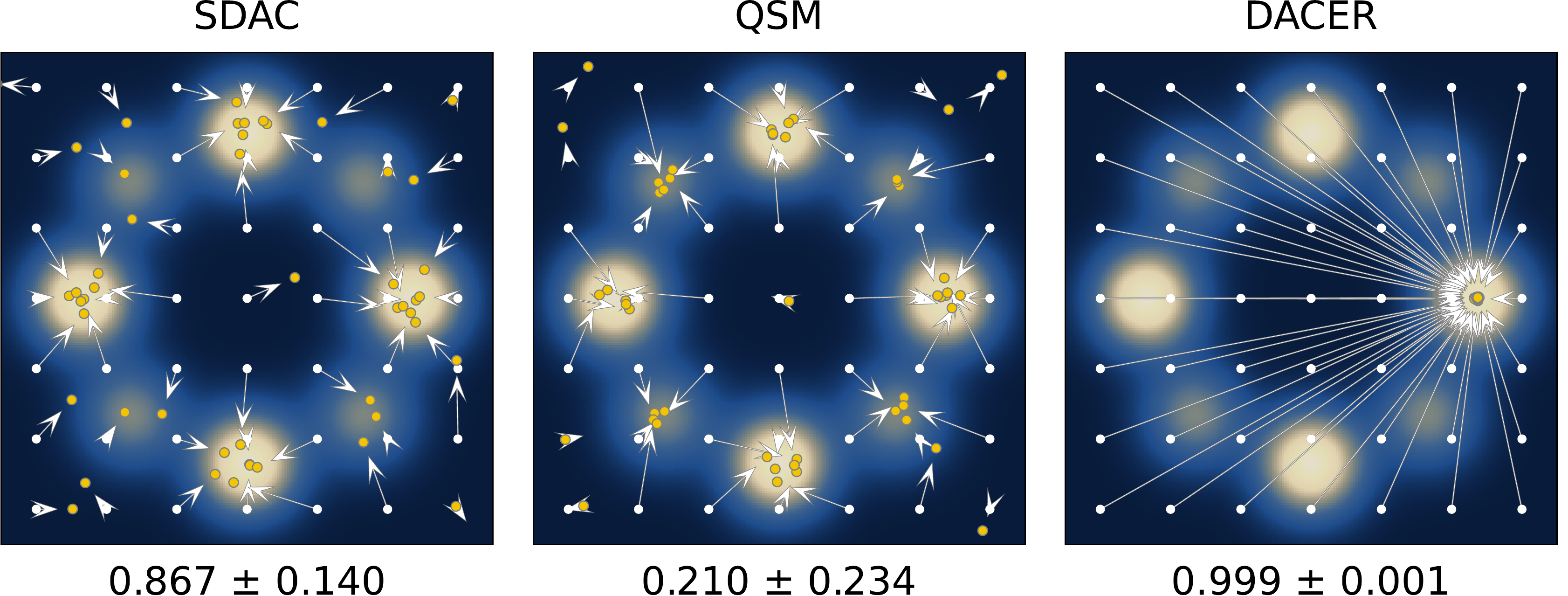}
    \includegraphics[width=0.0427\linewidth]{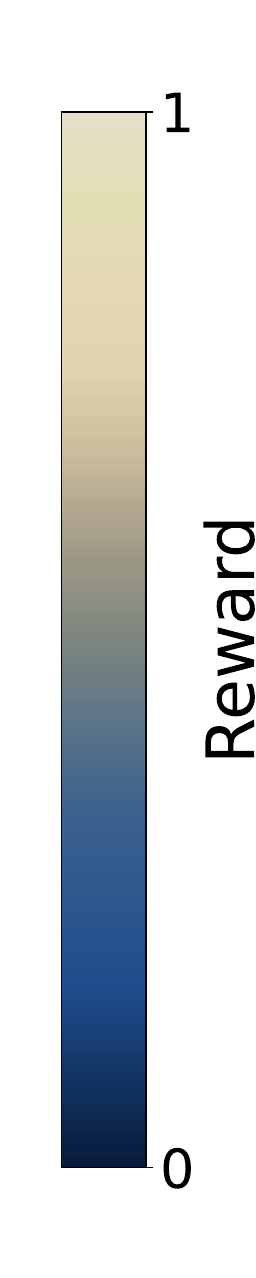}
    \caption{
    Visualization of samples obtained from different algorithms in the 2D bandit environment. White dots denote initial samples, and arrows indicate their corresponding denoised actions (yellow) sampled by each method. The numbers below each plot show the mean reward $\pm$ standard deviation, computed over 10k denoised samples. Detailed experimental setup is provided in Appendix~\ref{appendix:bandit_env}.}
    \label{fig:sampling_compare}
    \vspace{-0.2cm}
\end{figure*}

\begin{figure*}[t!]
    \centering
    \includegraphics[width=1.0\linewidth]{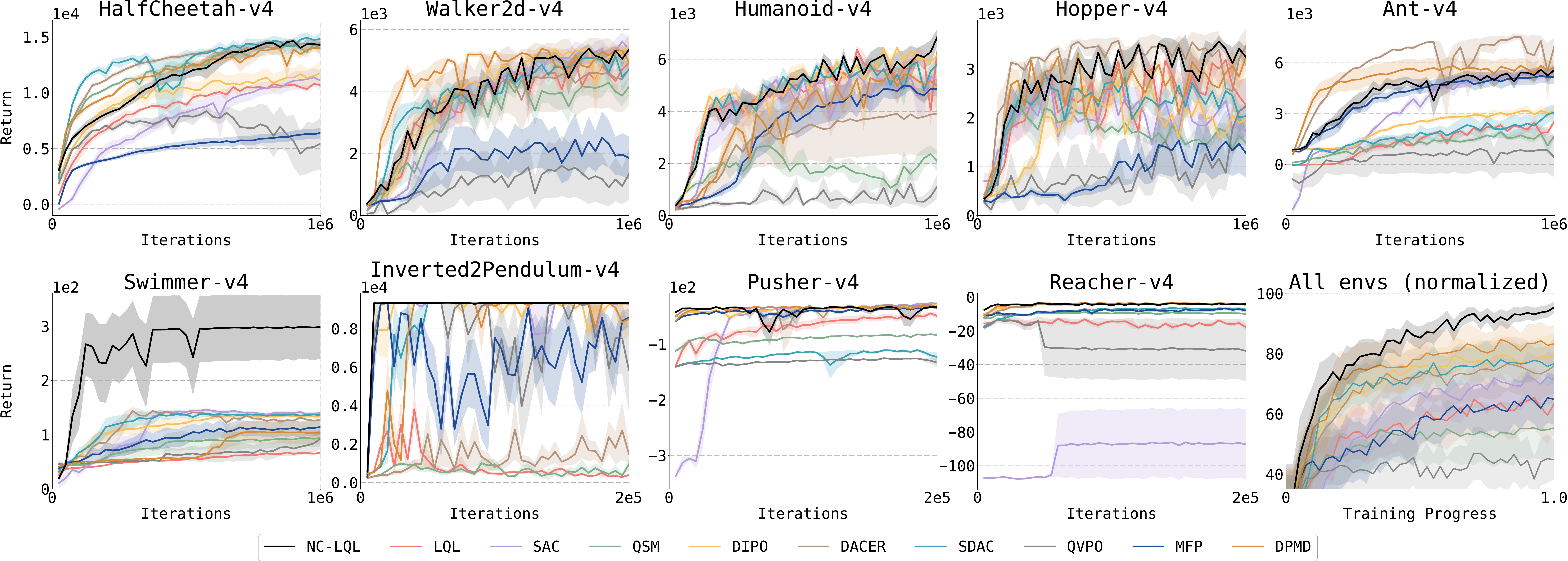}
    \caption{Training performance on OpenAI Gym MuJoCo environments. Each curve reports the mean return over 5 random seeds, with shaded regions indicating the standard error.}
    \label{fig:online_experiments}
    \vspace{-0.2cm}
\end{figure*}

\section{Experiments}

\subsection{2D Bandit Environment}
To further evaluate whether diffusion-based policies can reliably identify and concentrate on high-reward regions without being distracted by suboptimal local modes, we conduct experiments on the same 2D bandit environment across different baselines, as shown in~\cref{fig:sampling_compare}. DACER~\citep{wang2024diffusion} collapses entirely into a single mode, as it directly trains the diffusion model to maximize $Q$, which drives samples aggressively toward the highest-valued region and induces severe mode collapse. QSM~\citep{psenka2023learning} does not leverage a noise-conditioned value representation, leading to misaligned gradients during the denoising process and consequently insufficient coverage of high-value modes. SDAC~\citep{ma2025efficient} captures multimodality more effectively than other diffusion-based baselines by employing a carefully designed diffusion training objective, but still leaves a subset of samples trapped in local optima and incurs substantial algorithmic complexity. In contrast, NC-LQL successfully captures all four high-reward modes by sampling from progressively smoothed value landscapes. In Appendix~\ref{appendix:bandit_measure}, we provide a quantitative evaluation of multimodality in the 2D bandit environment.

\begin{table}[t!]
\centering
\caption{Normalized performance across all environments over training progress, reported as the mean return $\pm$ standard error over 5 seeds. At each progress, the \textcolor{red}{\textbf{best}} and \textbf{second-best} methods are highlighted in \textcolor{red}{\textbf{red}} and \textbf{bold}, respectively.}
\resizebox{\columnwidth}{!}{%
\begin{tabular}{lccccc}
\toprule
 Progress & 0.2 & 0.4 & 0.6 & 0.8 & 1.0 \\
\midrule
SAC   
& $46.1 {\scriptstyle \pm 9.8}$ & $63.9 {\scriptstyle \pm 8.0}$ & $66.4 {\scriptstyle \pm 7.8}$ & $70.7 {\scriptstyle \pm 0.4}$ & $71.3 {\scriptstyle \pm 8.7}$ \\
QSM   
& $45.6 {\scriptstyle \pm 9.1}$ & $53.3 {\scriptstyle \pm 9.1}$ & $55.0 {\scriptstyle \pm 9.8}$ & $53.9 {\scriptstyle \pm 0.3}$ & $55.3 {\scriptstyle \pm 9.6}$ \\
DIPO  
& $62.2 {\scriptstyle \pm 9.1}$ & 74.8 ${\scriptstyle \pm 6.9}$ & $75.7 {\scriptstyle \pm 6.3}$ & $79.6 {\scriptstyle \pm 6.8}$ & $79.8 {\scriptstyle \pm 6.9}$ \\
DACER 
& $61.0 {\scriptstyle \pm 10.5}$ & $71.9 {\scriptstyle \pm 9.6}$ & $69.4 {\scriptstyle \pm 9.3}$ & $75.5 {\scriptstyle \pm 9.7}$ & $76.0 {\scriptstyle \pm 9.5}$ \\
QVPO  
& $42.9 {\scriptstyle \pm 10.7}$ & $40.8 {\scriptstyle \pm 8.8}$ & $38.6 {\scriptstyle \pm 7.0}$ & $45.0 {\scriptstyle \pm 9.3}$ & $44.7 {\scriptstyle \pm 6.6}$ \\
MFP   
& $44.1 {\scriptstyle \pm 10.4}$ & $54.5 {\scriptstyle \pm 8.8}$ & $61.0 {\scriptstyle \pm 8.0}$ & $61.4 {\scriptstyle \pm 8.0}$ & $64.8 {\scriptstyle \pm 8.5}$ \\
SDAC  
& $65.1 {\scriptstyle \pm 6.3}$ & $69.6 {\scriptstyle \pm 6.1}$ & $73.3 {\scriptstyle \pm 6.6}$ & $78.0 {\scriptstyle \pm 6.5}$ & $76.7 {\scriptstyle \pm 6.5}$ \\
DPMD  
& \textcolor{red}{\textbf{72.0}} ${\scriptstyle \pm 9.2}$ & $\textbf{75.4} {\scriptstyle \pm 8.6}$ & \textbf{80.2} ${\scriptstyle \pm 7.6}$ & \textbf{84.7} ${\scriptstyle \pm 6.4}$ & \textbf{83.3} ${\scriptstyle \pm 6.3}$ \\
\midrule
\textbf{LQL}    
& $52.6 {\scriptstyle \pm 7.4}$ & $56.2 {\scriptstyle \pm 9.2}$ & $57.7 {\scriptstyle \pm 9.8}$ & $63.7 {\scriptstyle \pm 10.0}$ 
& $63.4 {\scriptstyle \pm 10.0}$ \\
\textbf{NC-LQL} 
& \textbf{71.2} ${\scriptstyle \pm 7.3}$ & \textcolor{red}{\textbf{84.5}} ${\scriptstyle \pm 4.1}$ & \textcolor{red}{\textbf{91.4}} ${\scriptstyle \pm 3.1}$ & \textcolor{red}{\textbf{92.2}} ${\scriptstyle \pm 2.6}$ & \textcolor{red}{\textbf{95.5}} ${\scriptstyle \pm 2.0}$ \\
\bottomrule
\end{tabular}}
\label{tab:online_experiments}
\vspace{-0.5cm}
\end{table}

\begin{figure*}
    \centering
    \includegraphics[width=1.0\linewidth]{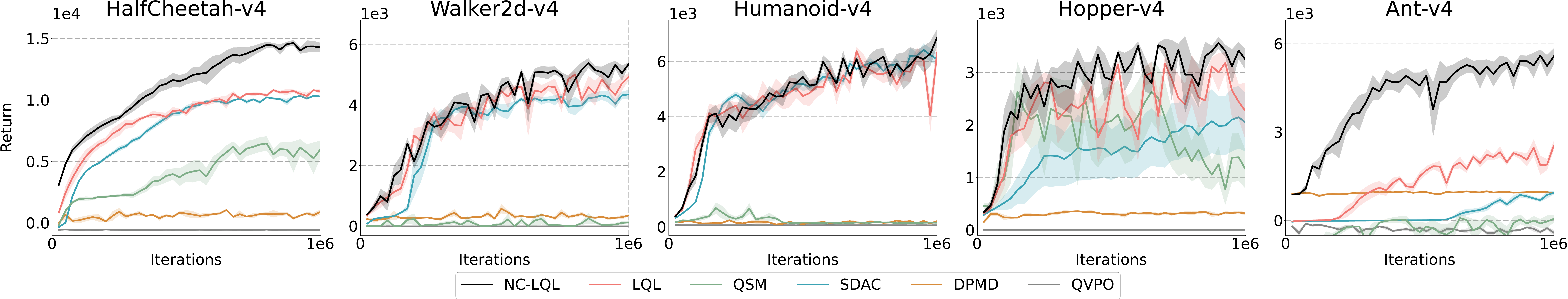}
    \caption{Training performance on OpenAI Gym MuJoCo environments using a single action sample (without best-of-$N$ sampling). Each curve shows the mean return over 5 random seeds, with shaded regions indicating the standard error.}
    \label{fig:best_of_N}
    \vspace{-0.2cm}
\end{figure*}

\subsection{OpenAI Gym MuJoCo Environment}
We evaluate the online RL performance of LQL and NC-LQL on a suite of OpenAI Gym MuJoCo control tasks. As baselines, we consider the standard off-policy actor-critic algorithm SAC~\citep{haarnoja2018soft} along with several diffusion-based actor-critic methods: QSM~\citep{psenka2023learning}, DIPO~\citep{yang2023policy}, QVPO~\citep{ding2024diffusion}, DACER~\citep{wang2024diffusion}, MFP~\citep{mean_flow_policy_2025} and SDAC, DPMD~\citep{ma2025efficient}.

\cref{fig:online_experiments} shows training curves for each of the 9 environments, along with an additional curve showing performance normalized across all environments. Algorithms are trained for 1M steps on 6 environments (\texttt{HalfCheetah-v4}, \texttt{Walker2d-v4}, \texttt{Humanoid-v4}, \texttt{Hopper-v4}, \texttt{Ant-v4}, and \texttt{Swimmer-v4}) and for 200k steps on 3 environments (\texttt{Pusher-v4}, \texttt{Reacher-v4}, and \texttt{InvertedDoublePendulum-v4}). Across MuJoCo benchmarks, NC-LQL demonstrates performance that is competitive with state-of-the-art diffusion-based baselines. In particular, when results are normalized and aggregated across all environments, NC-LQL consistently outperforms existing methods, indicating robust and stable improvements. By introducing multi-scale noise perturbations into the value function, NC-LQL effectively reshapes the sampling geometry, enabling efficient exploration across modes and reliable convergence to high-value actions. These results empirically validate that multi-scale noise conditioning provides a principled and effective solution to the slow-mixing issue inherent in LQL. Additional experimental details and ablation studies are provided in Appendix~\ref{appendix:exp_details}.

\paragraph{Normalized Performance} 
\Cref{tab:online_experiments} reports normalized performance aggregated across all environments at different training progress. Despite its simple and actor-free formulation, NC-LQL consistently achieves the highest normalized returns throughout training, substantially outperforming recent state-of-the-art diffusion-based algorithms such as DPMD~\cite{ma2025efficient}.

\paragraph{Single Action Sample Evaluation}
Best-of-$N$ sampling requires generating and evaluating multiple candidate actions per decision step, which substantially increases inference-time computation and limits practical deployment. For this reason, \cref{fig:best_of_N} reports training performance when actions are generated using a single action sample per decision step, without employing best-of-$N$ sampling. While NC-LQL natively operates without best-of-$N$ action selection, diffusion-based baselines commonly rely on this heuristic; in this experiment, we disable best-of-$N$ sampling for all methods to ensure a fair comparison. Under this restricted setting, NC-LQL consistently achieves strong performance across tasks, demonstrating that superior performance can be attained without relying on multiple candidate actions or diffusion-based policy parameterizations.

\begin{wraptable}{r}{0.37\linewidth}
    \centering
    \vspace{-0.4cm}
    \caption{Normalized parameter counts.}
    \vspace{-0.1cm}
    \resizebox{\linewidth}{!}{
        \begin{tabular}{lc}
        \toprule
        \textbf{Method} & \textbf{\# Params} \\
        \midrule
        SAC   & 1.477 \\
        DIPO  & 1.495 \\
        DACER & 1.489 \\
        QVPO  & 1.487 \\
        SDAC  & 1.487 \\
        DPMD  & 1.487 \\
        \midrule
        LQL & 0.978 \\  % 465410
        \textbf{NC-LQL} & \textbf{1.000} \\
        \bottomrule
        \end{tabular}
    }
    \label{tab:params}
    \vspace{-0.55cm}
\end{wraptable}
\paragraph{Model Complexity} We also evaluate the model complexity of our algorithms relative to baselines. \cref{tab:params} reports parameter counts in the \texttt{Humanoid-v4} environment, where NC-LQL requires substantially fewer parameters as a result of removing the actor network. NC-LQL uses only 475k parameters (normalized to $1.0$), which is substantially fewer than diffusion-based methods, and even fewer than SAC with 702k parameters ($1.477$). This lightweight design reduces architectural complexity while maintaining competitive performance, establishing NC-LQL as a practical alternative to previous online RL methods.

\begin{wraptable}{r}{0.35\linewidth}
    \centering
    \vspace{-0.48cm}
    \caption{Normalized Training Time.}
    \vspace{-0.1cm}
    \resizebox{\linewidth}{!}{
        \begin{tabular}{lc}
        \toprule
        \textbf{Method} & \textbf{Time} \\
        \midrule
        DIPO  & 1.505 \\
        DACER & 1.899 \\
        QVPO  & 5.515 \\
        SDAC  & 1.687 \\
        DPMD  & 1.343 \\
        \midrule
        LQL & 0.909 \\
        \textbf{NC-LQL} & \textbf{1.000} \\
        \bottomrule
        \end{tabular}
    }
    \label{tab:time}
    \vspace{-0.55cm}
\end{wraptable}
\paragraph{Wall-Clock Training Time}
We evaluate wall-clock training time on the \texttt{Humanoid-v4} environment using a single NVIDIA RTX 4090 GPU. As summarized in~\cref{tab:time}, NC-LQL completes training with a normalized time of 1.0, corresponding to 1.65 hours, while diffusion-based online RL baselines require between 1.34$\times$ and 5.52$\times$ longer training time. This gap arises because diffusion-based methods rely on explicit actor optimization and entropy or likelihood approximations, while NC-LQL eliminates these components altogether.

\section{Conclusion}
We introduced Langevin Q-Learning (LQL), an actor-free framework that directly samples actions from the soft policy using Langevin dynamics driven by the Q-function gradient. While conceptually simple and principled, LQL suffers from slow mixing in complex value landscapes. To address this limitation, we proposed Noise-Conditioned Langevin Q-Learning (NC-LQL), which incorporates multi-scale noise conditioning into the Q-function to enable efficient annealed sampling. Experiments on OpenAI Gym MuJoCo benchmarks show that NC-LQL achieves competitive or superior performance, while maintaining a simple and lightweight learning pipeline.

\paragraph{Limitations} NC-LQL currently lacks an automatic mechanism for adjusting the temperature $w$, which has to be tuned manually. Future work includes extending NC-LQL with soft Q-functions~\citep{haarnoja2017reinforcement, haarnoja2018soft} to better capture entropy-regularized objectives and developing adaptive strategies for automatic temperature tuning.

% In the unusual situation where you want a paper to appear in the
% references without citing it in the main text, use \nocite
% \nocite{langley00}

\bibliography{example_paper}
\bibliographystyle{icml2026}

%%%%%%%%%%%%%%%%%%%%%%%%%%%%%%%%%%%%%%%%%%%%%%%%%%%%%%%%%%%%%%%%%%%%%%%%%%%%%%%
%%%%%%%%%%%%%%%%%%%%%%%%%%%%%%%%%%%%%%%%%%%%%%%%%%%%%%%%%%%%%%%%%%%%%%%%%%%%%%%
% APPENDIX
%%%%%%%%%%%%%%%%%%%%%%%%%%%%%%%%%%%%%%%%%%%%%%%%%%%%%%%%%%%%%%%%%%%%%%%%%%%%%%%
%%%%%%%%%%%%%%%%%%%%%%%%%%%%%%%%%%%%%%%%%%%%%%%%%%%%%%%%%%%%%%%%%%%%%%%%%%%%%%%
\newpage
\appendix
\onecolumn
\section{Related Works}
\label{appendix:related_works}
\paragraph{Gaussian Homotopy Optimization}
Homotopy optimization~\cite{blake1987visual, allgower2012numerical, dunlavy2005homotopy}, also known as continuation optimization, is a general strategy for solving complicated non-convex optimization problems of the form $\min_{\mathbf{x} \in \mathbb{R}^d} f(\mathbf{x})$. This method introduces a continuation parameter $t \in [0, 1]$ and constructs a family of surrogate functions $H(\mathbf{x}, t)$, called a homotopy, which gradually transforms an easy-to-optimize function $g(\mathbf{x})$ into the target objective $f(\mathbf{x})$. Specifically, the homotopy satisfies $H(\mathbf{x}, 0) = g(\mathbf{x})$ and $H(\mathbf{x}, 1) = f(\mathbf{x})$. By tracing the solution path from the minimizer of $g(\mathbf{x})$ as $t$ increases from $0$ to $1$, the algorithm can effectively avoid poor local optima that typically hinder standard gradient-based methods.

Gaussian Homotopy (GH) is a specific instance of this framework that employs Gaussian smoothing to construct the surrogate functions~\cite{mobahi2015theoretical, iwakiri2022single}. In GH, the homotopy function is defined as the expectation of the objective function under Gaussian perturbations:
\begin{align}
\label{eq:gh}
    GH(\mathbf{x}, t) := \int f(\mathbf{x} + \beta(1-t)\epsilon) \mathcal{N}(\epsilon;\mathbf{0},\mathbf{I}) d\epsilon,
\end{align}
where $\beta > 0$ controls the smoothing range, $\epsilon$ denotes the noise vector sampled from a standard Gaussian distribution. As $t$ approaches $1$, the smoothing effect diminishes, recovering the original objective $f(\mathbf{x})$. This operation effectively widens the basins of attraction for the global optimum, facilitating better convergence in highly non-convex landscapes.

\paragraph{Denoising Diffusion Probabilistic Models} Diffusion models~\citep{sohl2015deep,ho2020denoising} are a class of generative models that construct samples through a Markov forward-reverse process. In the forward process, Gaussian noise is incrementally added to a clean data sample $\mathbf{x}_0$ over $T$ timesteps, according to $ q(\mathbf{x}_t|\mathbf{x}_{t-1}) := \mathcal{N}(\mathbf{x}_t; \sqrt{1-\beta_t}\mathbf{x}_{t-1}, \beta_t \mathbf{I})$, where $\beta_t \in (0,1)$ is a predefined variance schedule and $t$ denotes the diffusion timestep. Importantly, this process admits a closed-form expression for sampling $\mathbf{x}_t$ from $\mathbf{x}_0$ at any timestep $t$, $q(\mathbf{x}_t|\mathbf{x}_0) := \mathcal{N}(\mathbf{x}_t; \sqrt{\bar{\alpha}_t}\mathbf{x}_0, (1-\bar{\alpha}_t)\mathbf{I}),$
with $\alpha_t := 1-\beta_t$ and $\bar{\alpha}_t:=\prod^t_{s=1}\alpha_s$. The reverse process starts from standard Gaussian noise $\mathbf{x}_T \sim \mathcal{N}(\mathbf{0}, \mathbf{I})$ and progressively denoises through a parameterized Markov chain:
\begin{align}
\label{eq:diff_reverse}
    \mathbf{x}_{t-1} = \frac{1}{\sqrt{\alpha_t}} \left(\mathbf{x}_t - \frac{\beta_t}{\sqrt{1-\bar{\alpha}_t}} \epsilon_\theta(\mathbf{x}_t, t)\right) + \sigma_t \mathbf{z}, \quad \text{where} \quad \mathbf{z} \sim \mathcal{N}(\mathbf{0}, \mathbf{I}),
\end{align}
where the variance is fixed as $\sigma_t^2 = \frac{1- \bar{\alpha}_{t-1}}{1-\bar{\alpha}_t}\beta_t$. The diffusion model $\epsilon_\theta$ is trained to approximate the added noise by minimizing a simplified surrogate objective derived from a variational bound:
\begin{align}
\mathbb{E}_{\mathbf{x}_0 \sim \mathcal{B},\epsilon \sim \mathcal{N}(\mathbf{0}, \mathbf{I}),t\sim \mathcal{U}[1,T]}\left[\left\Vert \epsilon - \epsilon_\theta\left(\sqrt{\bar{\alpha}_t}\mathbf{x}_0 + \sqrt{1-\bar{\alpha}_t}\epsilon, t\right)\right\Vert^2\right].
\end{align}

\paragraph{Diffusion Models in Offline RL}
Diffusion models have recently been established as powerful policy representations in offline RL, providing a natural way to capture multimodal behaviors. \citet{wang2022diffusion} introduce conditional diffusion models that combine behavior cloning with Q-learning to achieve strong performance. \citet{ajay2022conditional} reformulates offline decision-making as conditional diffusion-based trajectory generation, leveraging iterative denoising for planning without explicit value functions. \citet{janner2022planning} propose trajectory-level denoising for planning, enabling long-horizon reasoning and flexible goal conditioning. \citet{chen2023score} presents a behavior-regularized policy optimization framework based on a pretrained diffusion behavior model. \citet{kang2023efficient} propose an efficient diffusion policy for offline reinforcement learning that avoids costly reverse diffusion chains during training while enabling compatibility with likelihood-based policy optimization methods. \citet{lu2023contrastive} formulate energy-guided sampling to realize principled Q-guided optimization. \citet{chen2024diffusion} introduces a diffusion trust region framework that distills expressive diffusion behavior policies into efficient one-step policies for offline reinforcement learning. \citet{frans2025diffusion} reinterpret diffusion model guidance as a controllable policy improvement operator, enabling test-time adjustment of optimality without explicit policy optimization. \citet{lu2025makes} conduct a large-scale empirical study of diffusion planning for offline reinforcement learning, systematically analyzing key design choices and proposing a simple yet strong diffusion planning baseline, and \citet{ki2025prior} propose Prior Guidance, a guided sampling framework for diffusion planners that learns a behavior-regularized prior to generate high-value trajectories without inference-time sample selection.

\paragraph{Diffusion Models in Online RL} 
In online RL, diffusion policies have been adapted to support continual interaction and efficient policy improvement. \citet{yang2023policy} establishes the first formulation of diffusion policies with convergence guarantees. \citet{psenka2023learning} propose Q-score matching, which trains diffusion model policies by directly matching the policy score to the action-gradient of the Q-function, enabling expressive and multimodal policy learning without backpropagating through the full diffusion process. \citet{ding2024diffusion} propose a variational lower bound on the policy objective, enabling sample-efficient online updates with entropy regularization. \citet{wang2024diffusion} treats the reverse process itself as the policy, introducing adaptive exploration control through entropy estimation. \citet{celik2025dime} formulate maximum entropy reinforcement learning with diffusion-based policies by deriving a tractable lower bound on the entropy objective, enabling principled exploration without adding external noise. Most recently, \citet{ma2025efficient} generalized denoising objectives to train policies directly on value-based targets, yielding efficient online algorithms. 

\paragraph{Langevin Dynamics in Online RL}
While prior work has explored Langevin dynamics as a mechanism for principled exploration and policy optimization in online reinforcement learning, these approaches primarily incorporate Langevin dynamics either at the level of value function posterior sampling or within diffusion-based policy updates. \citet{ishfaq2023provable} propose Langevin Monte Carlo Least-Squares Value Iteration, which induces principled exploration by approximately sampling Q-functions from their posterior via Langevin dynamics. \citet{ishfaq2025langevin} proposed the Langevin Soft Actor-Critic (LSAC) algorithm, which achieves efficient exploration in continuous control tasks by leveraging distributional Langevin Monte Carlo (LMC) for approximate Thompson sampling and employing parallel tempering to effectively sample from multimodal Q-function posteriors. \citet{wang2025enhanced} leverage Langevin dynamics within diffusion-based policies to perform policy improvement. In contrast to these methods, NC-LQL directly realizes soft-policy sampling through annealed Langevin dynamics driven by a noise-conditioned Q-function, which mitigates the slow-mixing behavior of Langevin dynamics while eliminating the need for explicit actor optimization.

\section{Additional Pseudo-Code}
\subsection{Langevin Dynamics}
\begin{algorithm}[h!]
\caption{Langevin Dynamics}
\label{alg:ld}
\begin{algorithmic}
\STATE {\bfseries Input:} score network $s_\theta (\mathbf{x})$, total number of iterations $T$, step size $\epsilon$, initial sample $\mathbf{x}_{0}$
\FOR{$t \leftarrow 1$ to $T$}
\STATE Sample $\mathbf{z}_t \sim \mathcal{N}(\mathbf{0}, \mathbf{I})$
\STATE $\mathbf{x}_{t} \leftarrow \mathbf{x}_{t-1} + \frac{\epsilon}{2}s_\theta(\mathbf{x}_{t-1}) + \sqrt{\epsilon}\mathbf{z}_t$
\ENDFOR
\STATE \textbf{Return} $\mathbf{x}_{T}$ 
\end{algorithmic}
% \vspace{-0.5cm}
\end{algorithm}

\subsection{Langevin Soft Policy}
\begin{algorithm}[h!]
\caption{Langevin Soft Polciy}
\label{alg:lsp}
\begin{algorithmic}
\STATE {\bfseries Input:} value function $Q (\mathbf{s}, \mathbf{a})$, total number of iterations $T$, step size $\epsilon$, initial sample $\mathbf{a}_{0}$
\FOR{$t \leftarrow 1$ to $T$}
\STATE Sample $\mathbf{z}_t \sim \mathcal{N}(\mathbf{0}, \mathbf{I})$
\STATE $\mathbf{a}_{t} \leftarrow \mathbf{a}_{t-1} + \frac{\epsilon}{2}\nabla_{\mathbf{a}}Q(\mathbf{s}, \mathbf{a}_{t-1}) + \sqrt{\epsilon}\mathbf{z}_t$
\ENDFOR
\STATE \textbf{Return} $\mathbf{a}_{T}$ 
\end{algorithmic}
% \vspace{-0.5cm}
\end{algorithm}

\newpage
\subsection{Langevin Q-Learning}
\label{appendix:lql}

\begin{algorithm}[h!]
\caption{Langevin Q-Learning}
\begin{algorithmic}
\STATE {\bfseries Input:} replay buffer $\mathcal{B}$, value network $Q_\theta(\mathbf{s},\mathbf{a})$, total number of iterations $T$, step size $\epsilon$
\FOR{each iteration}
\FOR{each sampling step}
\STATE Sample $\mathbf{a} \sim \pi_{\text{LD}}(\cdot|\mathbf{s})$ by \cref{def:langevin_improvement}
\STATE Execute $\mathbf{a}$, observe reward $r$ and next state $\mathbf{s}'$
\STATE Store transition $(\mathbf{s}, \mathbf{a}, r, \mathbf{s}')$ in buffer $\mathcal{B}$
\ENDFOR
\FOR{each update step}
\STATE Sample mini-batch from $\mathcal{B}$
\STATE Update critic $Q_\theta$ with \cref{eq:lql_objective}
\ENDFOR
\ENDFOR
\end{algorithmic}
\end{algorithm}

\subsection{Noise-Conditioned Langevin Soft Policy}
\label{appendix:noise-conditioned langevin soft policy}
\begin{algorithm}[h!]
\caption{Noise-Conditioned Langevin Soft Policy}
\begin{algorithmic}
\STATE {\bfseries Input:} noise-conditioned value network $Q_{\text{NC}}(\mathbf{s},\tilde{\mathbf{a}},\sigma_i)$, total number of iterations $T$, step size $\epsilon$, noise scales $\{\sigma_i\}^L_{i=1}$, initial sample $\mathbf{a}_0$
\STATE $\mathbf{a}_{1,0} \leftarrow \mathbf{a}_0$
\FOR{$i \leftarrow 1$ to $L$}
\STATE $\alpha_i \leftarrow \epsilon \cdot \sigma^2_i / \sigma^2_L$
\FOR{$t \leftarrow 1$ to $T$}
\STATE Sample $\mathbf{z}_t \sim \mathcal{N}(\mathbf{0}, \mathbf{I})$
\STATE $\mathbf{a}_{i,t} \leftarrow \mathbf{a}_{i,t-1} + \frac{\alpha_i}{2} \nabla_{\tilde{\mathbf{a}}}Q_{\text{NC}}(\mathbf{s},\mathbf{a}_{i, t-1}, \sigma_i) + \sqrt{\alpha_i}\mathbf{z}_t$
\ENDFOR
\STATE $\mathbf{a}_{i+1,0} \leftarrow \mathbf{a}_{i,T}$
\ENDFOR
\STATE \textbf{Return} $\mathbf{a}_{L,T}$
\end{algorithmic}
\end{algorithm}

\newpage 

\section{Implementation Details in 2D Bandit Environments}
\label{appendix:bandit}

\subsection{Environment Settings}
\label{appendix:bandit_env}

\begin{wrapfigure}{r}{0.25\linewidth}
    \vspace{-0.5cm}
    \includegraphics[width=1\linewidth]{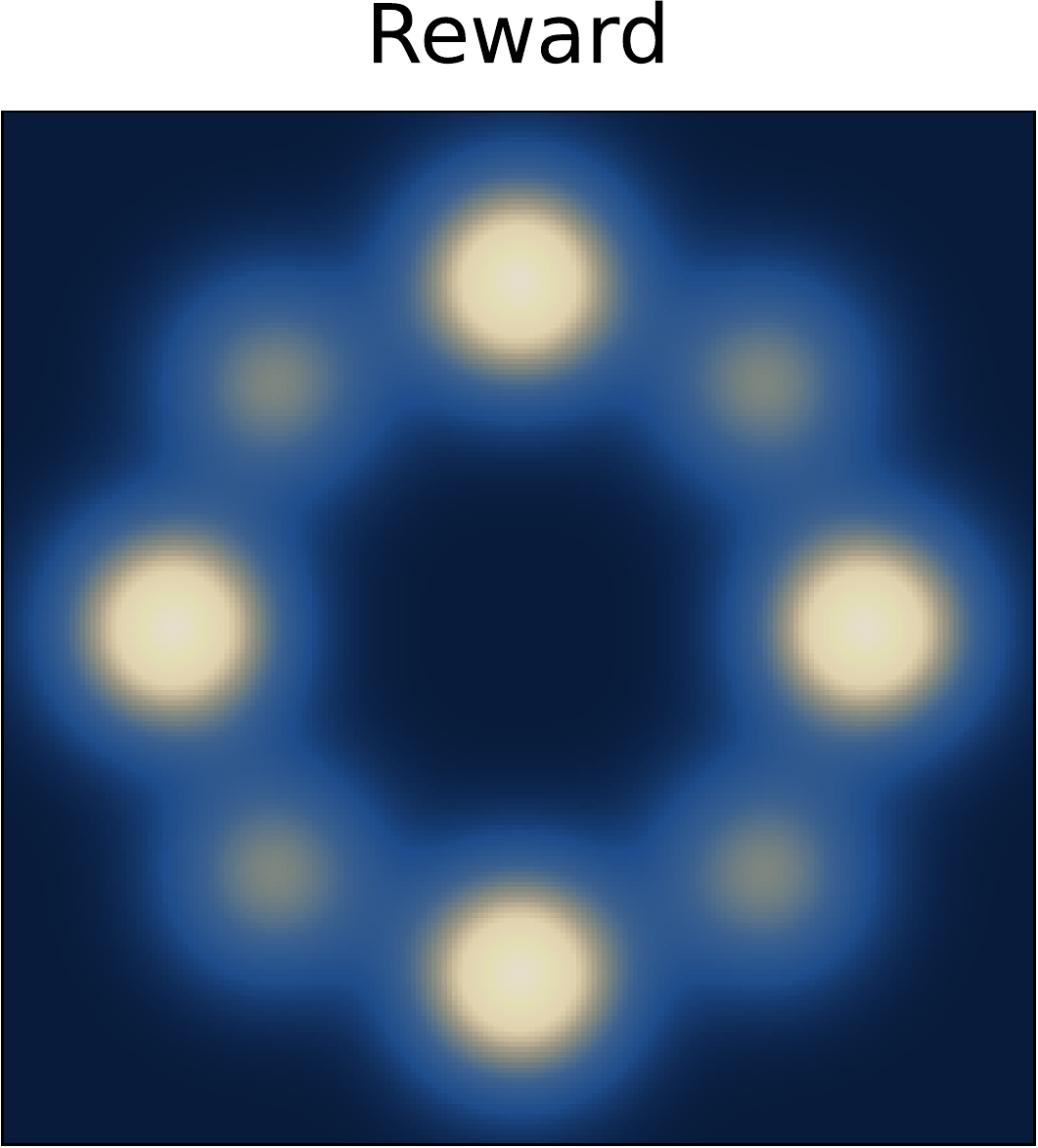}
    \caption{Reward map.}
    \label{fig:reward_map_2d_bandit}
    \vspace{-0.5cm}
\end{wrapfigure}
We design a multimodal reward function based on a mixture of Gaussian distributions, as illustrated in \cref{fig:reward_map_2d_bandit}. The reward corresponds to the probability density of this mixture, resulting in a landscape with eight modes, each represented by an isotropic Gaussian with covariance $0.3^2\mathbf{I}$. To induce asymmetry, alternating weights of 2 and 1 are assigned to the modes, which are positioned on a circle of radius $\sqrt{2}$ at coordinates $[(\sqrt{2},0),$ $(1,1),$ $(0,\sqrt{2}),$ $(-1,1),$ $(-\sqrt{2},0),$ $(-1,-1),$ $(0,-\sqrt{2}),$ $(1,-1)]$. This arrangement produces alternating high- and low-reward regions around the circle. The reward values are normalized so that the maximum equals 1.0. This structure highlights how NC-LQL’s smooth value function helps avoid convergence to local optima by effectively navigating multiple reward modes.

\paragraph{Figure~\ref{fig:sampling_compare}}
We select the temperature parameter $w$ for LQL and NC-LQL by measuring the average reward over 10k samples for $w \in [1, 500]$. The optimal values obtained from this sweep are used in the evaluations and action sampling.

\subsection{Quantitative Evaluation of Multi-Modality}
\label{appendix:bandit_measure}

\begin{table}[h!]
    \centering
    \caption{Proportion of samples reaching each high-value mode.}
    \begin{tabular}{lccccc}
    \toprule
    \textbf{Method} & \multicolumn{4}{c}{\textbf{Proportions}} & \textbf{Sum} \\
    \midrule
    SDAC  & 0.227 & 0.227 & 0.239 & 0.234 & 0.927 \\
    QSM   & 0.118 & 0.115 & 0.117 & 0.115 & 0.465 \\
    DACER & 0.000 & 1.000 & 0.000 & 0.000 & 1.000 \\
    DIPO  & 0.100 & 0.099 & 0.104 & 0.098 & 0.401 \\
    \midrule
    LQL & 0.141 & 0.147 & 0.139 & 0.143 & 0.570 \\
    \textbf{NC-LQL} & 0.240 & 0.256 & 0.243 & 0.254 & 0.993 \\
    \bottomrule
    \end{tabular}
    \label{tab:bandit-mm}
\end{table}

In \cref{tab:bandit-mm}, we report the proportion of samples assigned to each high-value mode for all methods. Sampling starts by drawing $\mathbf{a}_0 \sim \mathcal{N}(\mathbf{0}, \mathbf{I})$ and propagating through each algorithm’s sampling process. The four proportions correspond, in order from left to right, to the top, right, bottom, and left high-value modes. Each proportion is computed as the ratio of samples lying within an $\mathcal{L}_2$-distance of 0.3 from the mode center to the total number of samples (10k). Unlike the baselines, NC-LQL samples actions from a Boltzmann distribution via annealed Langevin dynamics driven by a noise-conditioned Q-function, which mitigates the slow-mixing behavior inherent to Langevin dynamics. As a result, NC-LQL achieves an aggregate score of 0.993 while maintaining nearly uniform sample proportions across all four modes ($\approx 0.25$).

\newpage
\section{Full Visualizations on 2D Bandit Environment}
\label{appendix:bandit_full}
To illustrate the intermediate denoising process in \cref{fig:sampling_compare}, we visualize samples at each denoising step. In this setting, initial actions are sampled from a uniform grid rather than from the standard normal distribution, and the number of denoising steps is fixed to 10 for all baselines. For NC-LQL, we set $T=1$ and $L=10$, while for LQL, we set $T=10$.
\begin{figure}[!htbp]
    \centering
    \includegraphics[scale=0.2]{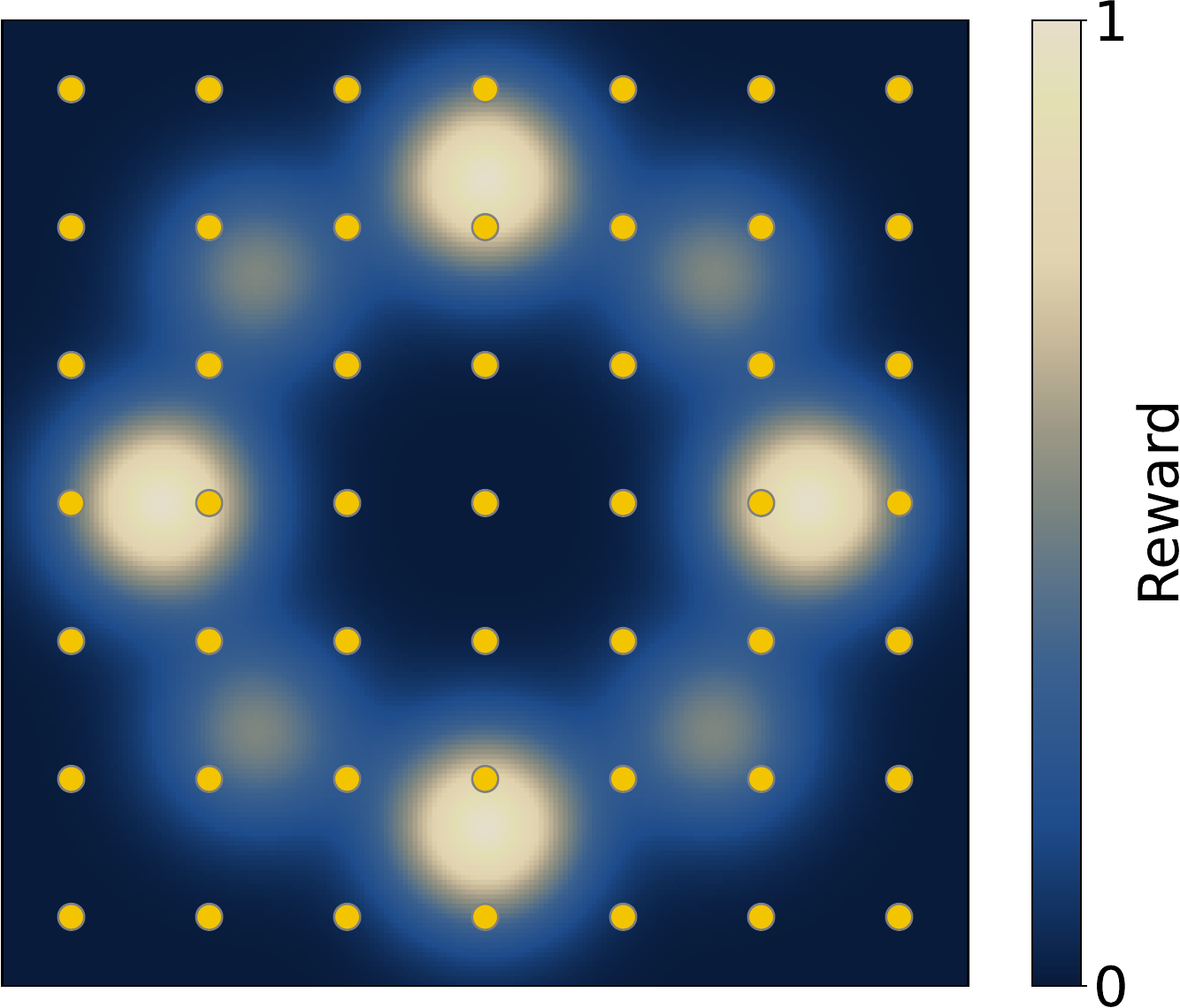}
    \caption{Initial samples.}
    \vspace{-0.3cm}
\end{figure}
 
\begin{figure}[!htbp]
    \centering
    \includegraphics[scale=0.2]{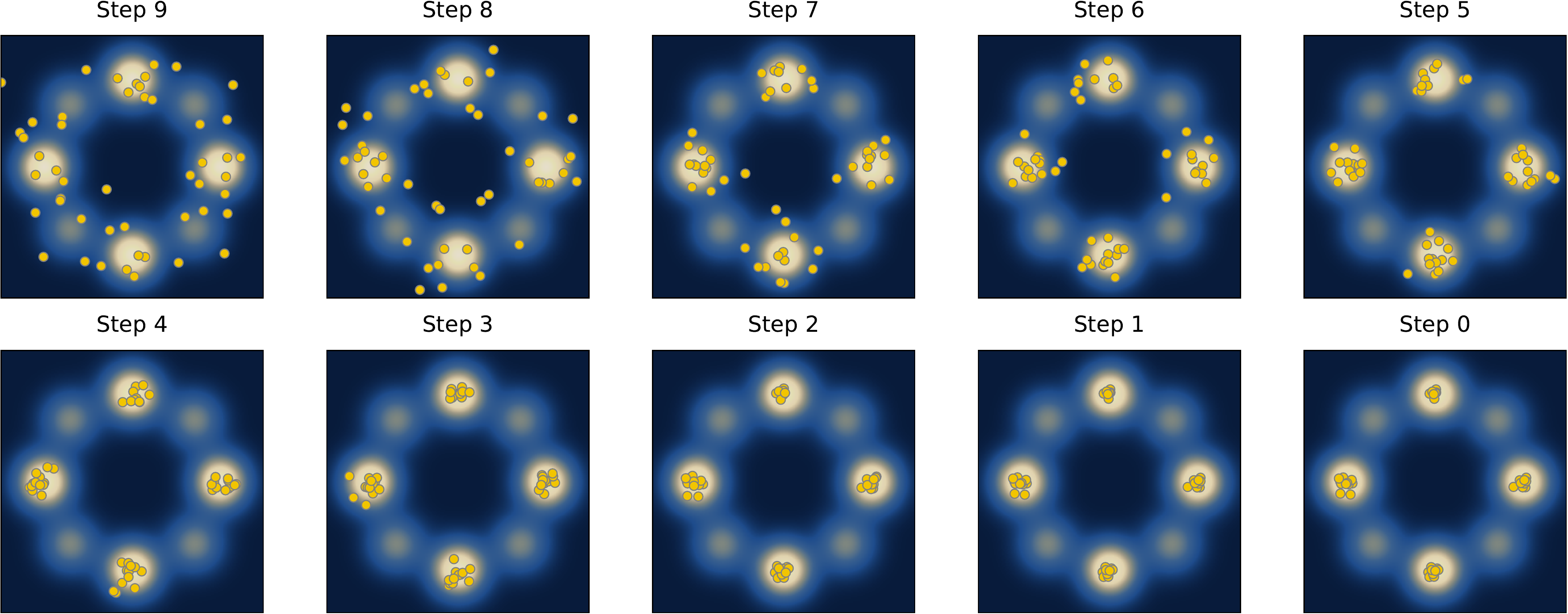}
    \caption{Visualization of intermediate denoising steps for NC-LQL.}
    \vspace{-0.3cm}
\end{figure}
 
\begin{figure}[!htbp]
    \centering
    \includegraphics[scale=0.2]{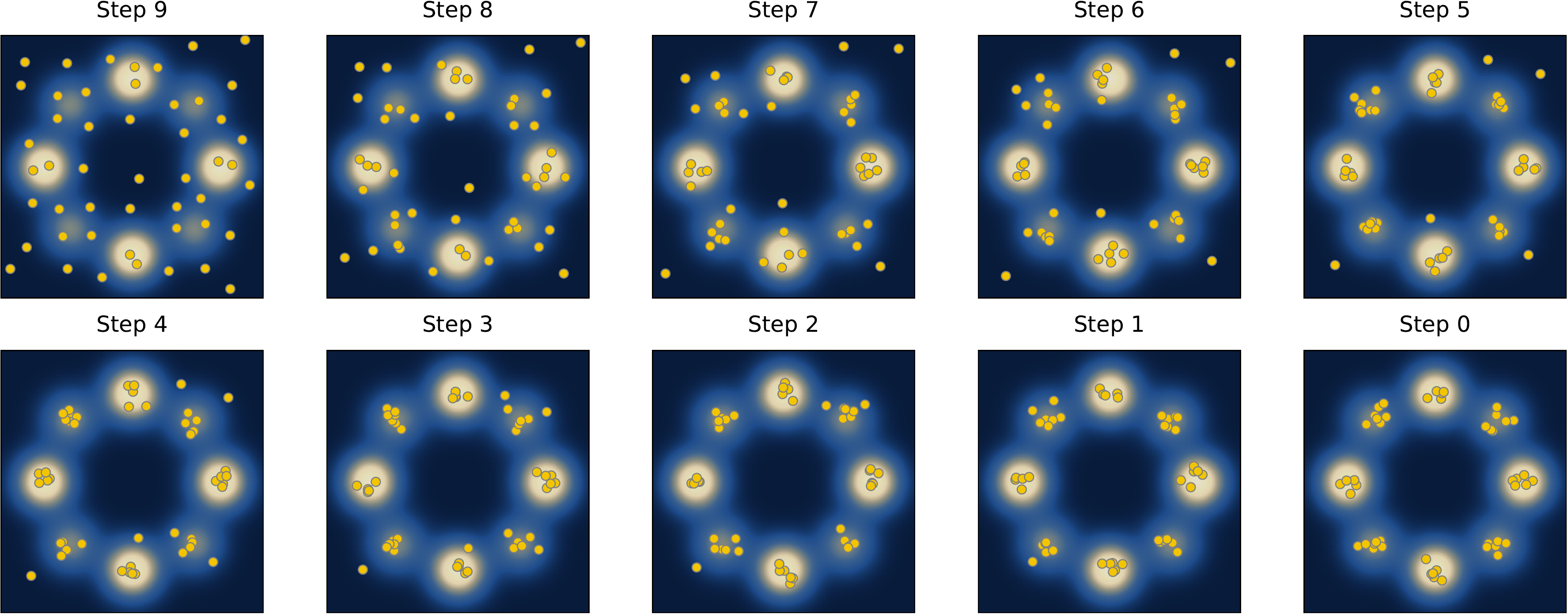}
    \caption{Visualizations of intermediate denoising steps for LQL.}
    \vspace{-0.3cm}
\end{figure}

\begin{figure}[!htbp]
    \centering
    \includegraphics[scale=0.2]{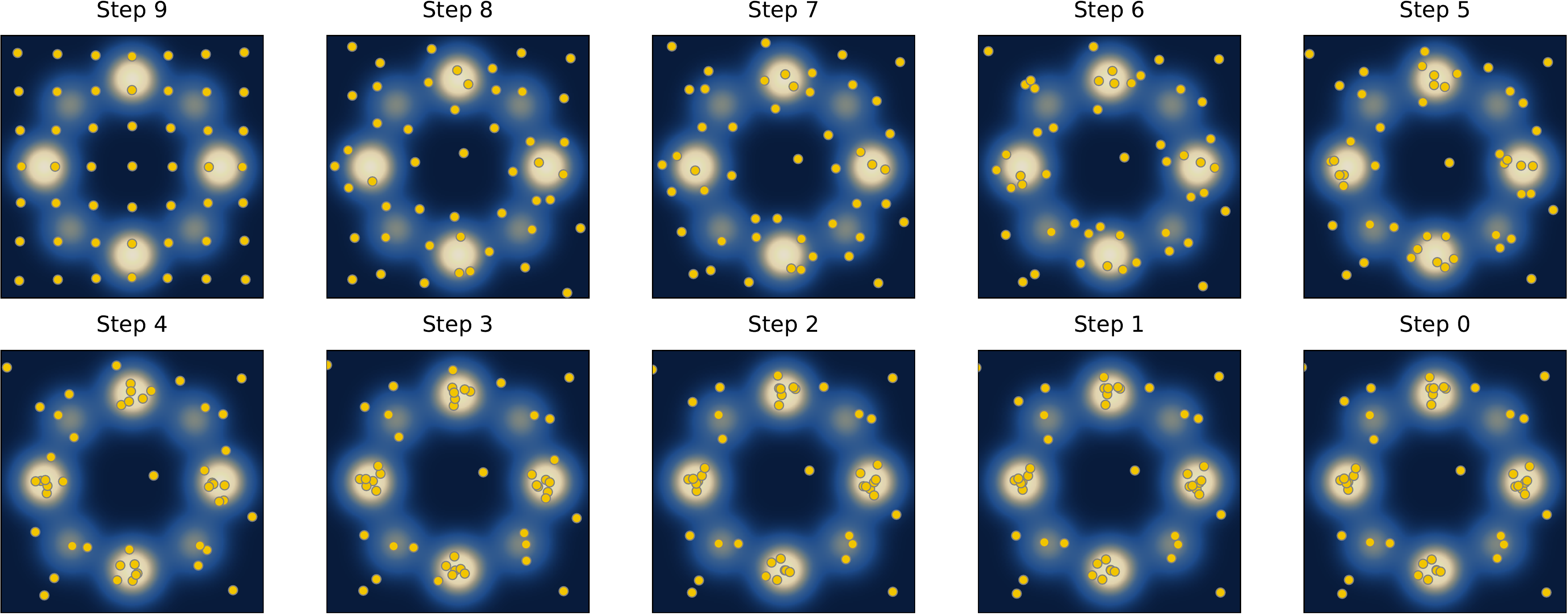}
    \caption{Visualizations of intermediate denoising steps for SDAC.}
    \vspace{-0.3cm}
\end{figure}
 
\begin{figure}[!htbp]
    \centering
    \includegraphics[scale=0.2]{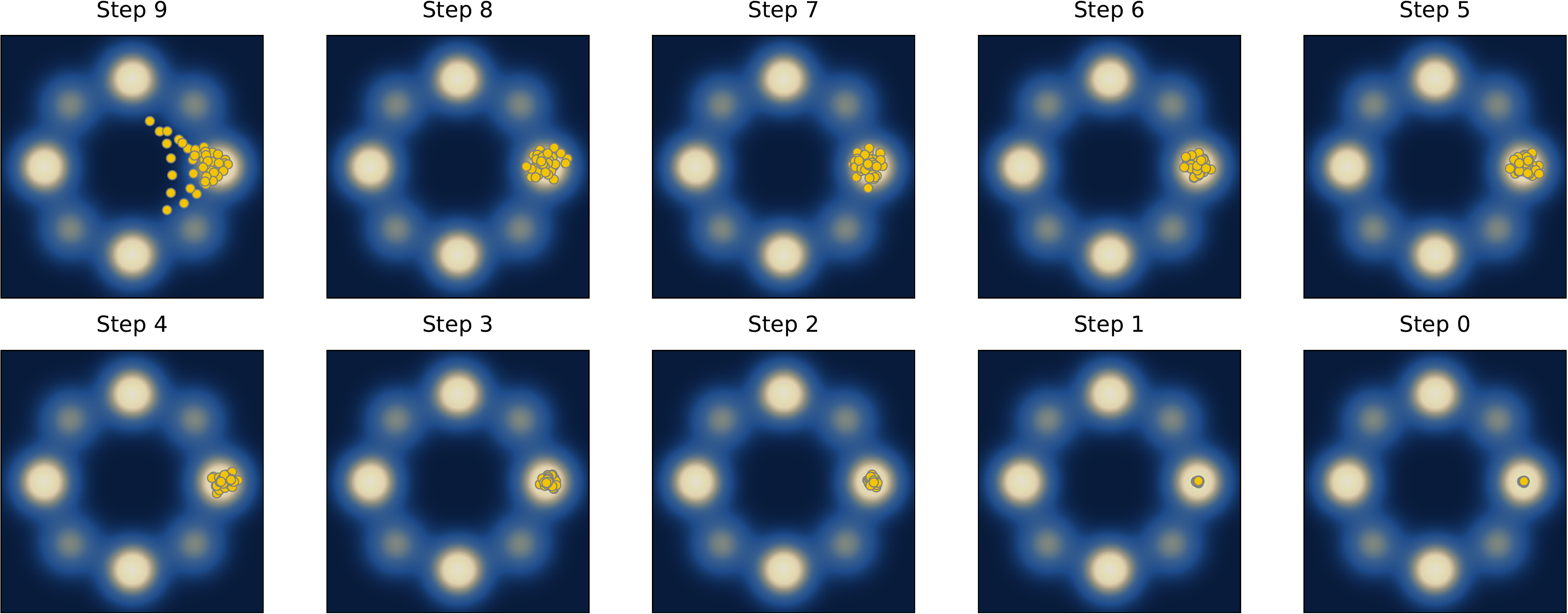}
    \caption{Visualizations of intermediate denoising steps for DACER.}
    \vspace{-0.3cm}
\end{figure}
 
\begin{figure}[!htbp]
    \centering
    \includegraphics[scale=0.2]{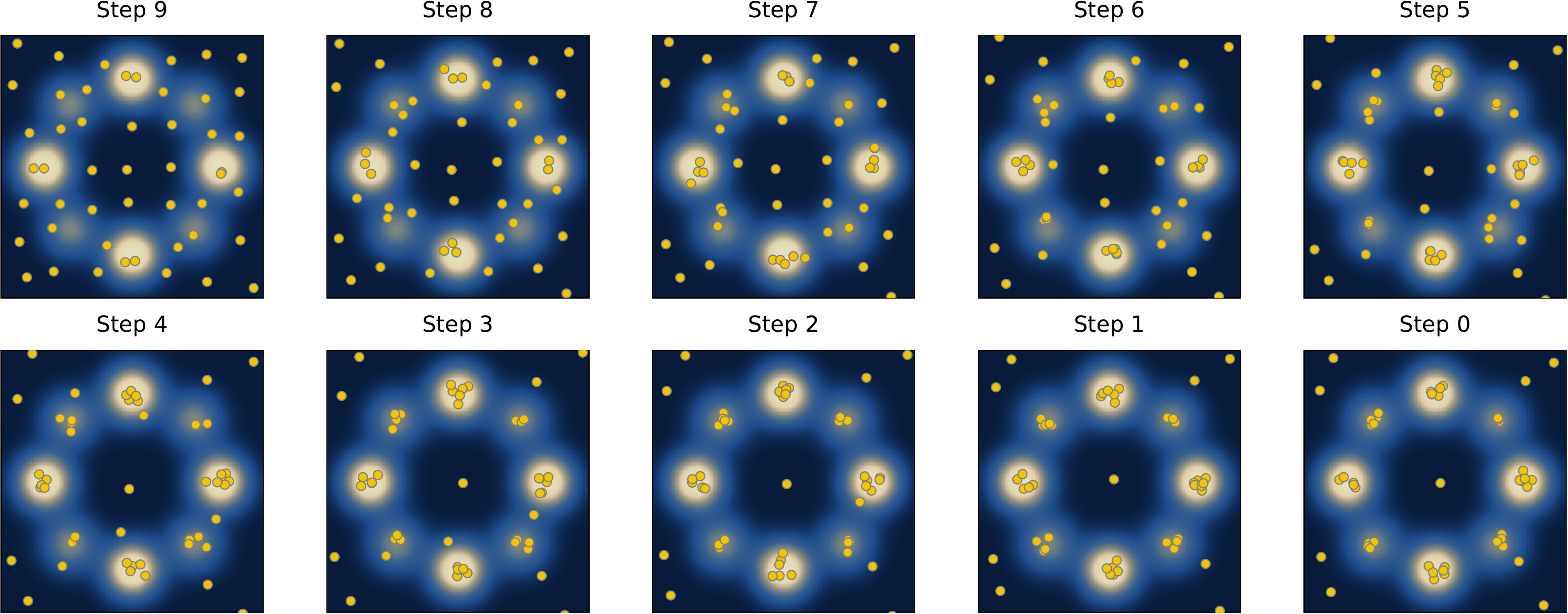}
    \caption{Visualizations of intermediate denoising steps for QSM.}
    \vspace{-0.3cm}
\end{figure}
 
\begin{figure}[!htbp]
    \centering
    \includegraphics[scale=0.2]{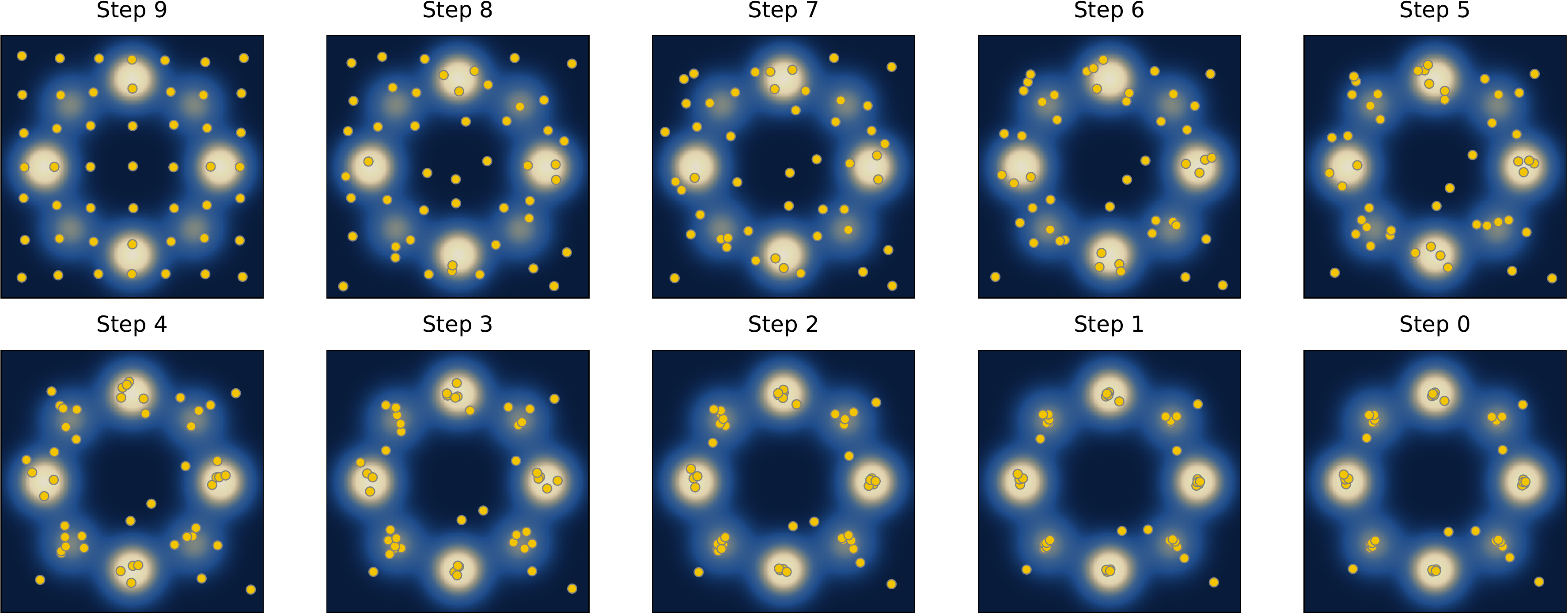}
    \caption{Visualizations of intermediate denoising steps for DIPO.}
    \vspace{-0.3cm}
\end{figure}

\newpage
\section{Experimental Details and Ablation Studies}
\label{appendix:exp_details}

\subsection{Hyperparameter Settings in Gym MuJoCo Environments}
Following \citet{ma2025efficient}, we collected data using five parallel vectorized environments for each task. Accordingly, the 1M training iterations reported in \cref{fig:online_experiments} correspond to a total of 5M environment interactions. The complete hyperparameter configurations for all baseline algorithms are provided in \cref{tab:hyperparams_baselines}.

For NC-LQL, we replace the diffusion timestep hyperparameter used in diffusion-based baselines with the total number of iterations $T$ and the number of noise steps $L$. To ensure a fair comparison, the total number of denoising iterations is fixed to $T \times L = 20$, matching that of the diffusion baselines. Detailed hyperparameter settings in LQL and NC-LQL are in~\cref{tab:detailed_hyperparameter}.

\begin{table}[h!]
    \centering
    \caption{The overall algorithms' hyperparameter settings.}
    \resizebox{\linewidth}{!}{%
    \begin{tabular}{lcccccccccc}
    \toprule
    \textbf{Hyperparameter}  &\textbf{LQL} &\textbf{NC-LQL} & \textbf{SDAC} & \textbf{DPMD}  & \textbf{QSM} & \textbf{DIPO} & \textbf{DACER} & \textbf{QVPO} & \textbf{SAC} &\textbf{MFP} \\ 
    \midrule
    Replay buffer capacity     &1e6  &1e6  & 1e6            & 1e6  & 1e6         & 1e6           & 1e6            & 1e6           & 1e6  & 1e6\\
    Buffer warm-up size        &3e4  &3e4   & 3e4            & 3e4  & 3e4          & 3e4           & 3e4            & 3e4           & 3e4 & 3e4 \\
    Batch size                 &256  &256   & 256            & 256  & 256         & 256           & 256            & 256           & 256 & 256 \\
    Discount factor $\gamma$   &0.99 &0.99   & 0.99           & 0.99  & 0.99        & 0.99          & 0.99           & 0.99          & 0.99 & 0.99\\
    Target update rate $\tau$  &0.005 &0.005   & 0.005        & 0.005 & 0.005        & 0.005         & 0.005          & 0.005         & 0.005 & 0.005\\
    Reward scale               &0.2 &0.2   & 0.2            & 0.2 & 0.2          & 0.2           & 0.2            & 0.2           & 0.2 & 0.2 \\
    No. of hidden layers       &3 &3     & 3              & 3 & 3            & 3             & 3              & 3             & 3 & 3 \\
    No. of hidden nodes        &256 &256   & 256            & 256 & 256          & 256           & 256            & 256           & 256 & 256 \\
    Activations                &Mish &Mish   & Mish      & Mish     & ReLU         & Mish          & Mish           & Mish          & GELU & GELU \\
    Diffusion steps            &- &-   & 20        & 20     & 20           & 100           & 20             & 20            & -  & 20\\
    total number of iterations $T$ & 20 & 2   & -    & -         & -           & -           & -             & -            & - & - \\
    noise steps $L$            & - & 10   & -      & -       & -           & -           & -             & -            & - & - \\ 
    Alpha delay update         &- &-  & 250       & 250     & -          & -           & 10,000         & -           & - & - \\
    Action gradient steps      &- &-   & -      & -      & -          & 30            & -            & -           & - & - \\
    No. of Gaussian distributions  &- &- & -   & -         & -          & -           & 3              & -           & - & - \\
    No. of action samples          &- &- & -     & -       & -          & -           & 200            & -           & - & - \\
    Noise scale       &- &-  & 0.1   & 0.1         & -          & -           & -            & -           & - & - \\
    Optimizer              &Adam &Adam       & Adam & Adam           & Adam         & Adam          & Adam           & Adam          & Adam & Adam \\
    Actor learning rate      &- &-     & 3e-4  & 3e-4          & 3e-4         & 3e-4          & 3e-4           & 3e-4          & 3e-4 &1e-4\\
    Critic learning rate       &1e-4 &1e-4   & 3e-4  & 3e-4          & 3e-4         & 3e-4          & 3e-4           & 3e-4          & 3e-4 &1e-4\\
    Alpha learning rate         &- &-  & 7e-3 & 7e-3           & -          & -           & 3e-2           & -           & 3e-4 &-\\
    Target entropy             &- &-   & -0.9 $\cdot$ dim($\mathcal{A}$) & -0.9 $\cdot$ dim($\mathcal{A}$) & - & - & -0.9 $\cdot$ dim($\mathcal{A}$) & - & -dim($\mathcal{A}$) & - \\
    No. of batch action sampling  &1 &1 & 32 & 32            & 32           & 1           & 1            & 32            & 1 & 32 \\ 
    \bottomrule
    \end{tabular}%
    }
    \label{tab:hyperparams_baselines}
\end{table}

\begin{table}[h!]
    \centering
    \caption{Detailed hyperparameter settings in LQL and NC-LQL.}
    \begin{tabular}{ccc}
        \toprule
        \textbf{Hyperparameter} & \textbf{LQL} & \textbf{NC-LQL}  \\
        \midrule
        Temperature parameter $w$ & 500 & 500 \\
        Step size $\epsilon$ &0.0001 &0.0001 \\
        Maximum noise level $\sigma_1$ &- &0.1 \\
        Minimum noise level $\sigma_L$ &- &0.001 \\
        \bottomrule
    \end{tabular}
    \label{tab:detailed_hyperparameter}
\end{table}

\subsection{Additional Implementation Details}
\paragraph{Score Normalization in Langevin Updates}
To further stabilize training, we normalize the score function during Langevin updates:
\begin{align*}
\nabla_{\tilde{\mathbf{a}}} Q_\theta(\mathbf{s}, \tilde{\mathbf{a}}, \sigma_i)
\leftarrow \nabla_{\tilde{\mathbf{a}}} Q_\theta(\mathbf{s}, \tilde{\mathbf{a}}, \sigma_i) \big/ \left(\Vert \nabla_{\tilde{\mathbf{a}}} Q_\theta(\mathbf{s}, \tilde{\mathbf{a}}, \sigma_i) \Vert + 1e^{-8} \right).
\end{align*}
This normalization prevents excessively large or uneven gradient magnitudes, which could otherwise lead to unstable updates. By ensuring a consistent scale, the denoising dynamics remain stable, and the critic can learn smoother value estimates. We apply this normalization to both LQL and NC-LQL.

\newpage
\subsection{Ablation Studies}
\label{appendix:ablation_studies}

\begin{figure}[h]
    \centering
    \includegraphics[width=1.0\linewidth]{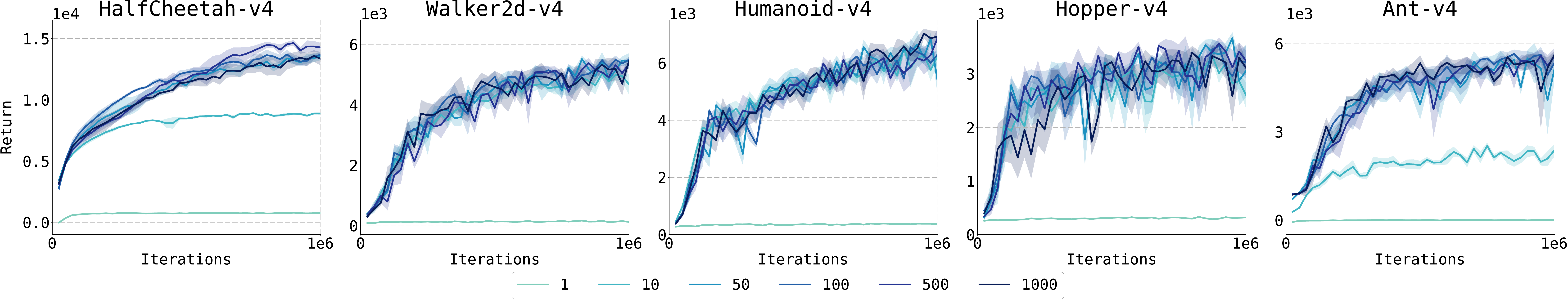}
    \caption{Training performance of NC-LQL across MuJoCo environments under varying temperature parameter $w$.  Each curve shows the mean return over 3 random seeds, with shaded regions indicating the standard error.}
    \vspace{-0.1cm}
\end{figure}
\begin{figure}[h]
    \centering
    \includegraphics[width=1.0\linewidth]{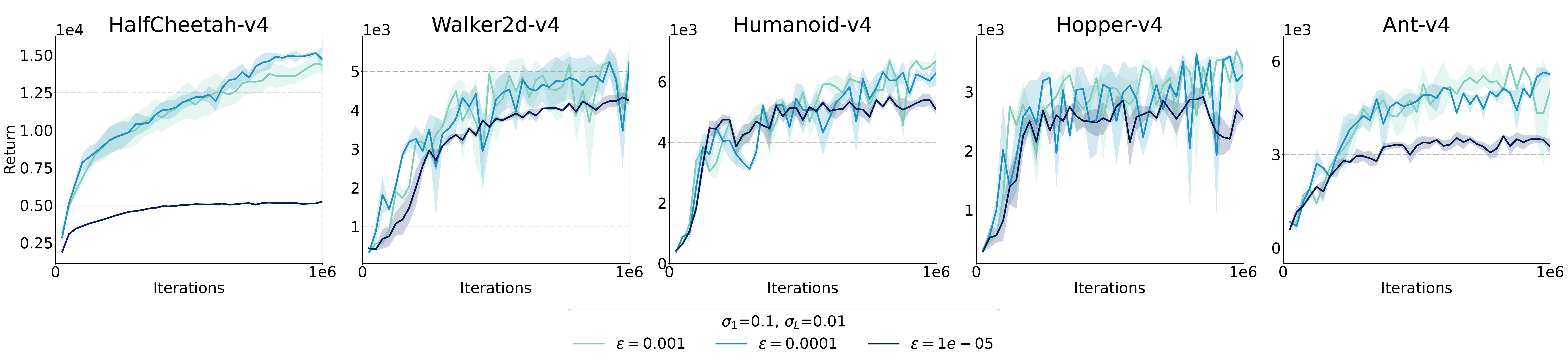}
    \includegraphics[width=1.0\linewidth]{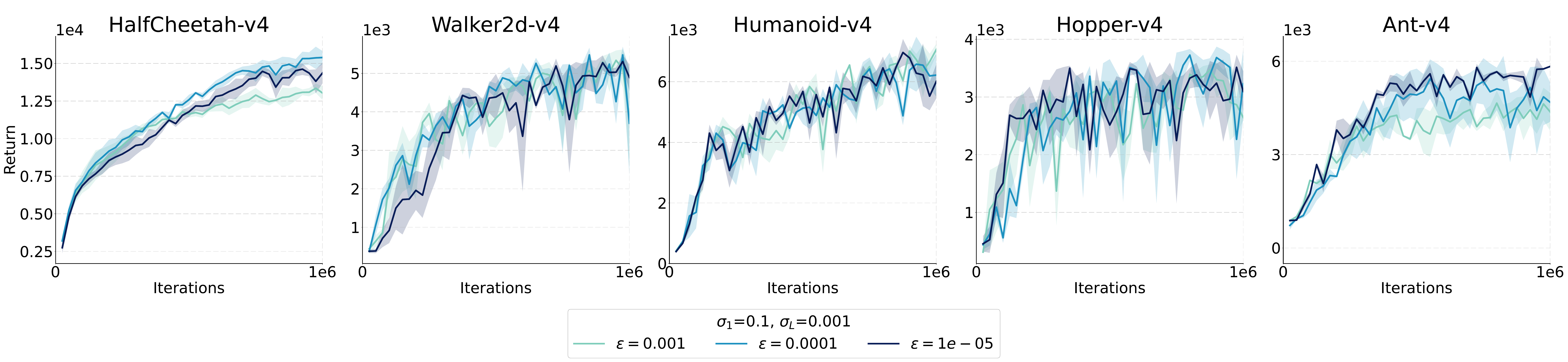}
    \includegraphics[width=1.0\linewidth]{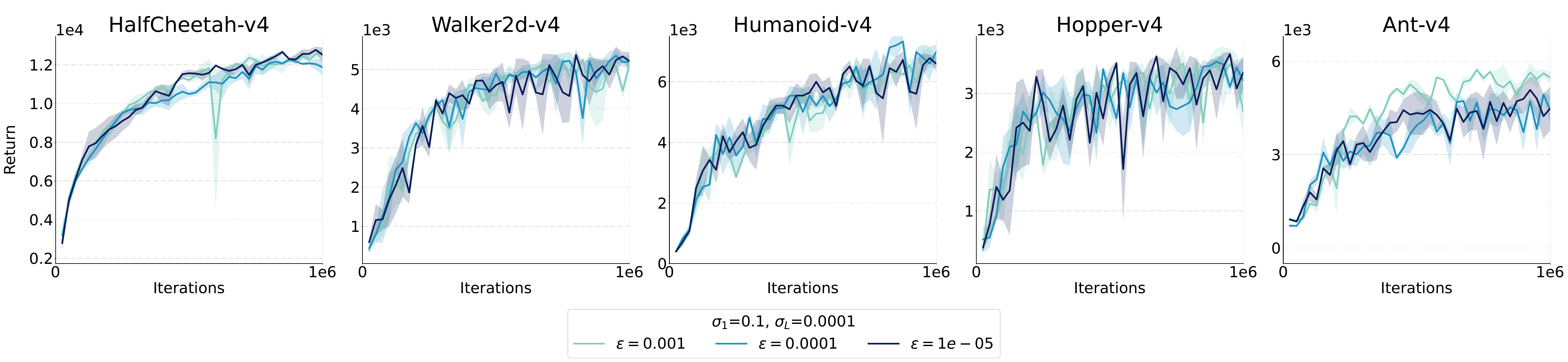}
    \caption{Training performance of NC-LQL across MuJoCo environments under varying temperature parameters $\sigma_L$ and $\epsilon$, with $\sigma_1$ fixed at 0.1. Each curve shows the mean return over 3 random seeds, with shaded regions indicating the standard error.}
\end{figure}
\begin{figure}[h]
    \centering
    \includegraphics[width=1.0\linewidth]{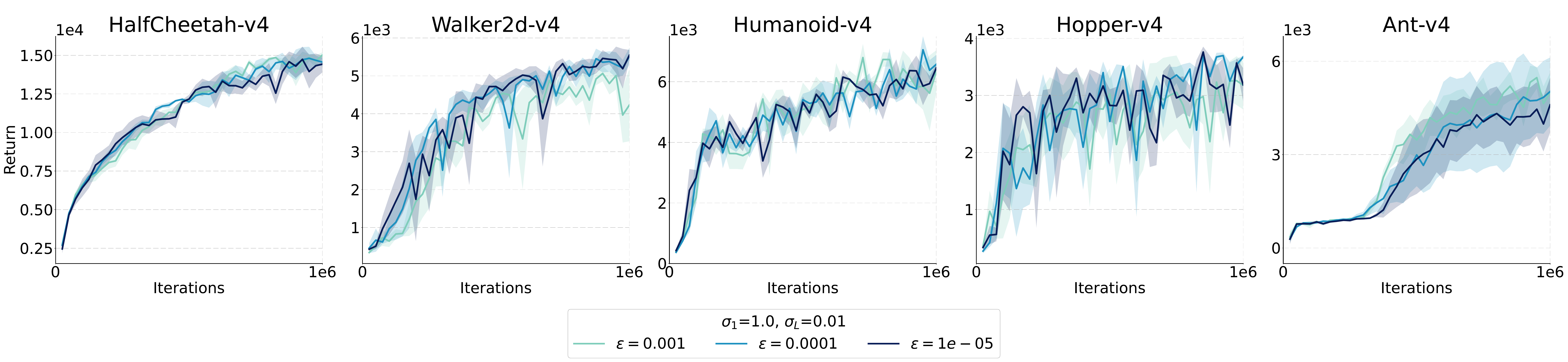}
    \includegraphics[width=1.0\linewidth]{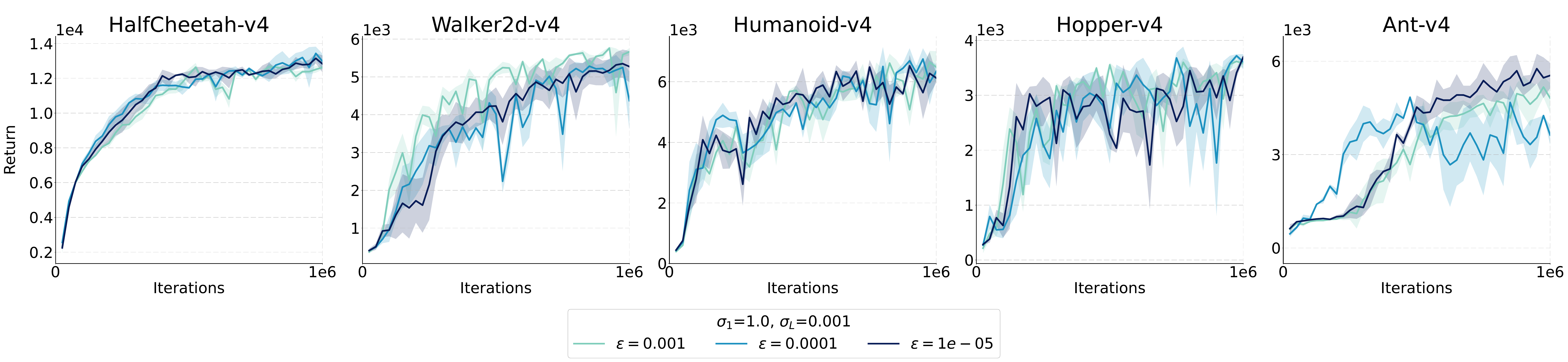}
    \includegraphics[width=1.0\linewidth]{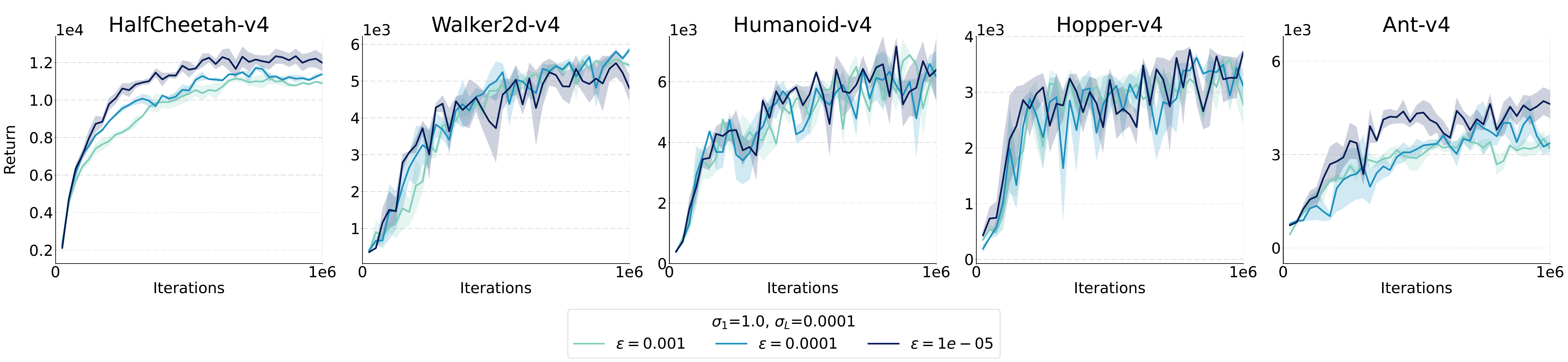}
    \caption{Training performance of NC-LQL across MuJoCo environments under varying temperature parameters $\sigma_L$ and $\epsilon$, with $\sigma_1$ fixed at 1.0. Each curve shows the mean return over 3 random seeds, with shaded regions indicating the standard error.}
\end{figure}
\begin{figure}[h]
    \centering
    \includegraphics[width=1.0\linewidth]{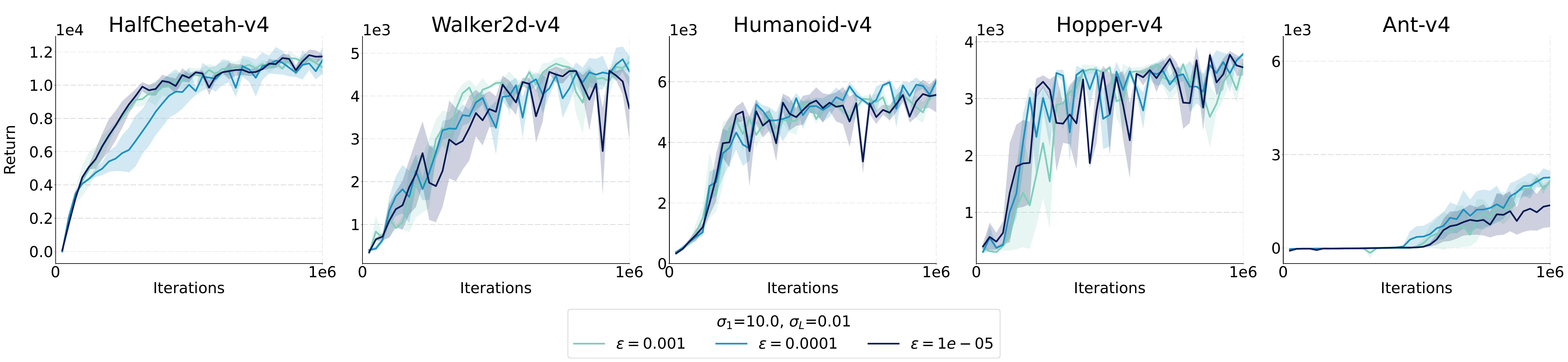}
    \includegraphics[width=1.0\linewidth]{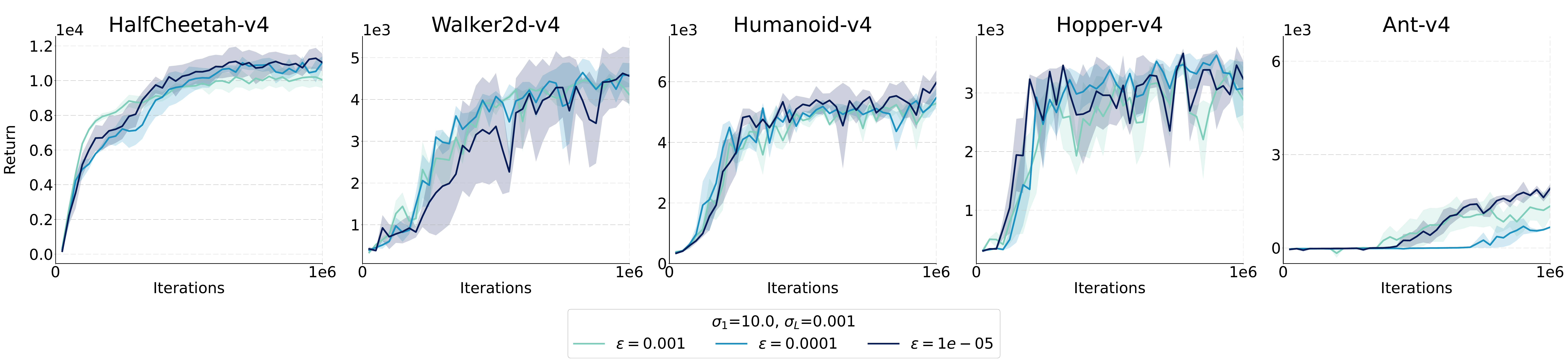}
    \includegraphics[width=1.0\linewidth]{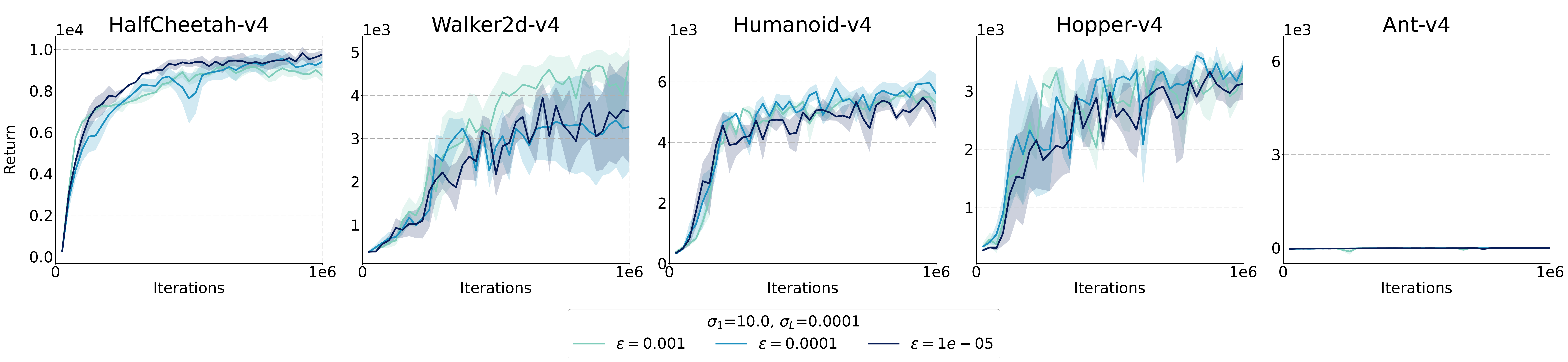}
    \caption{Training performance of NC-LQL across MuJoCo environments under varying temperature parameters $\sigma_L$ and $\epsilon$, with $\sigma_1$ fixed at 10.0. Each curve shows the mean return over 3 random seeds, with shaded regions indicating the standard error.}
\end{figure}

\begin{figure}[h]
    \centering
    \includegraphics[width=1.0\linewidth]{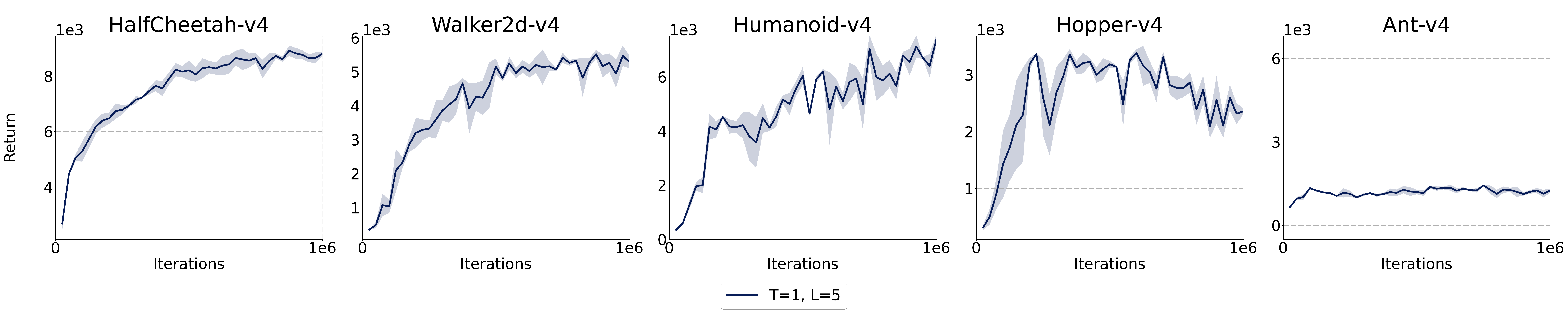}
    \includegraphics[width=1.0\linewidth]{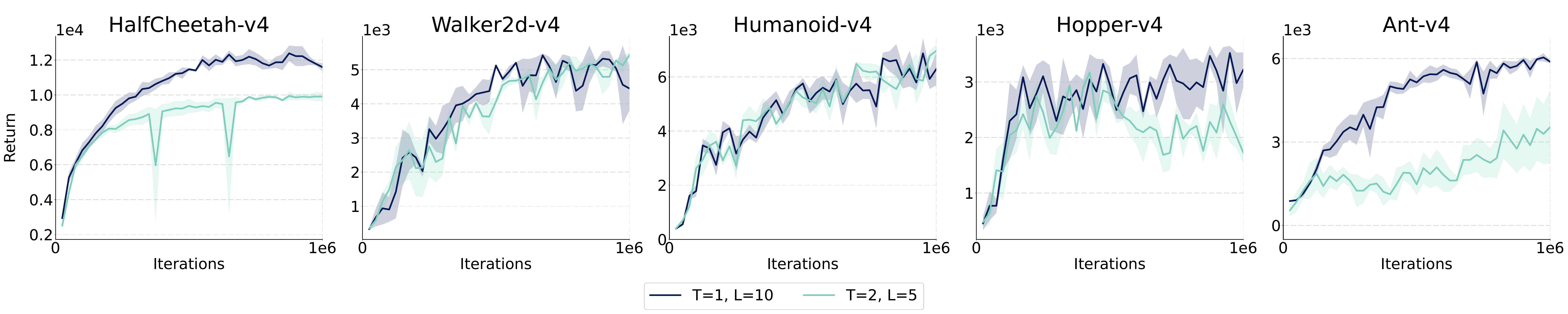}
    \includegraphics[width=1.0\linewidth]{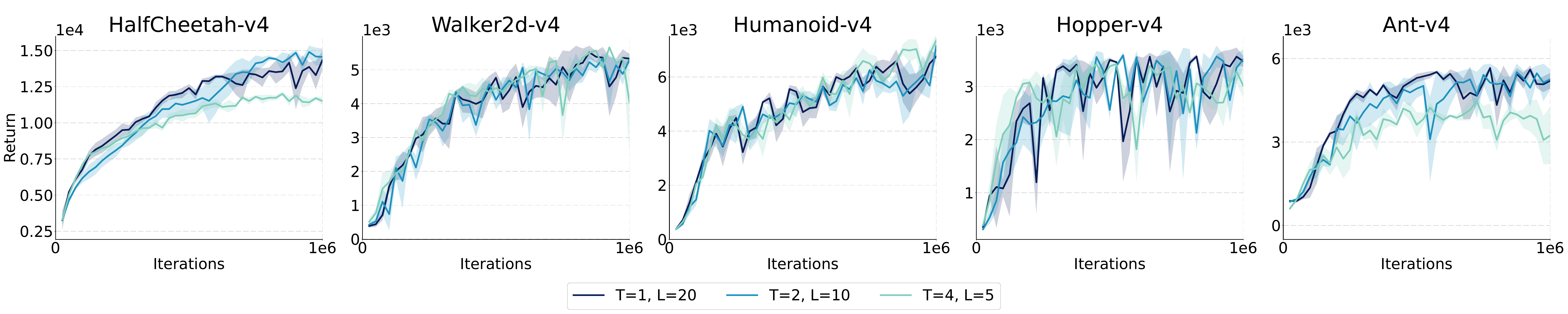}
    \includegraphics[width=1.0\linewidth]{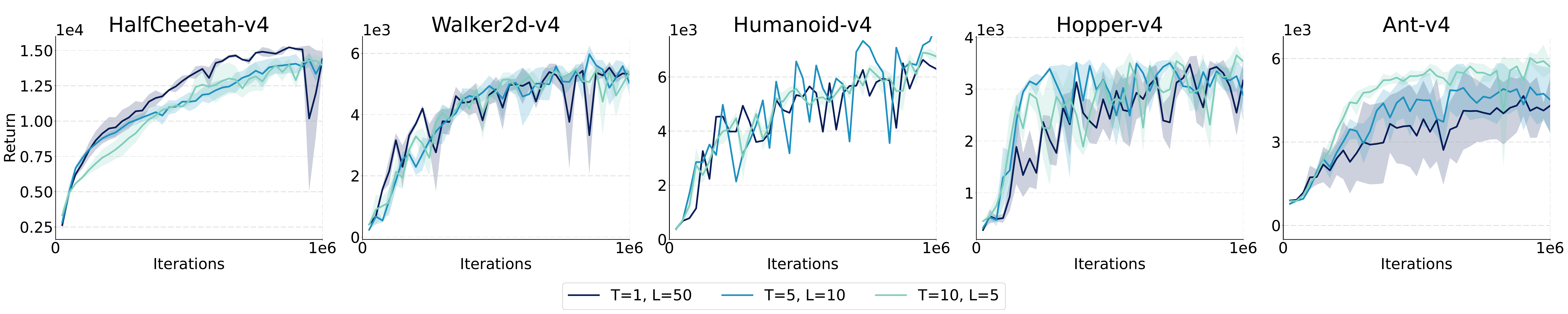}
    \includegraphics[width=1.0\linewidth]{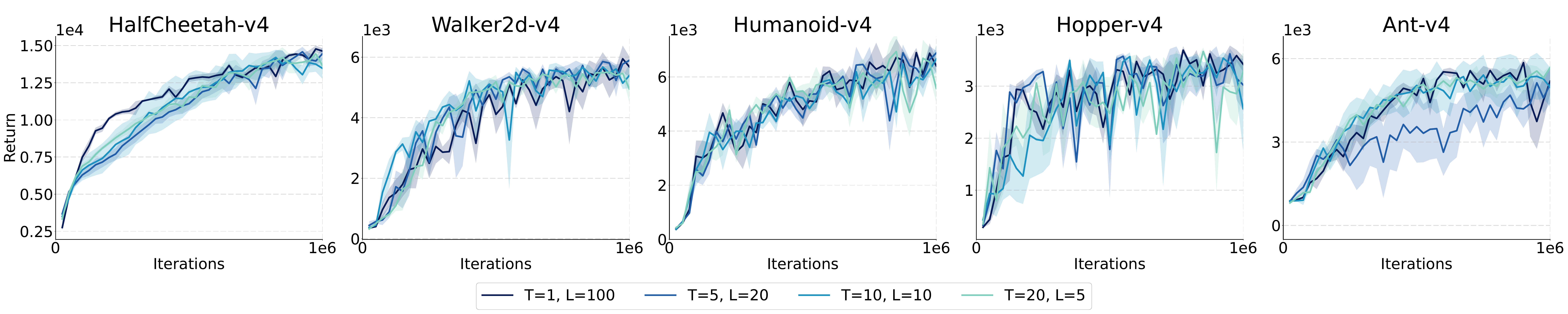}
    \caption{Training performance of NC-LQL on MuJoCo environments under varying total numbers of iterations $T$ and noise steps $L$. Each panel corresponds to a fixed product $T \times L$ (5, 10, 20, 50, and 100). Within each panel, we sweep different combinations of $T$ and $L$ that satisfy the same budget. Each curve shows the mean return over 3 random seeds, with shaded regions indicating the standard error.}
\end{figure}

%%%%%%%%%%%%%%%%%%%%%%%%%%%%%%%%%%%%%%%%%%%%%%%%%%%%%%%%%%%%%%%%%%%%%%%%%%%%%%%
%%%%%%%%%%%%%%%%%%%%%%%%%%%%%%%%%%%%%%%%%%%%%%%%%%%%%%%%%%%%%%%%%%%%%%%%%%%%%%%

\end{document}